\documentclass[twocolumn,twoside,a4paper]{article}
\usepackage[utf8]{inputenc}

\usepackage[
    paper=a4paper,
    top=1cm,
    bottom=1cm,
    left=1.5cm,
    right=1.5cm,
    headheight=1cm,
    footskip=0.2cm,
    headsep=0.2cm,
]{geometry} % 设置纸张和页面
\usepackage{fancyhdr} % 设置页眉页脚
\fancypagestyle{myheaderfooter}{%
    \fancyhf{} % 清除所有默认页眉和页脚
    \fancyhead[RO]{\thepage} % 在奇数页的右上角放置页码
    \fancyhead[LE]{\thepage} % 在偶数页的左上角放置页码
    \fancyhead[C]{Preprint version (TPAMI 2024) with the \href{https://github.com/lartpang/LaTeX-Templates}{template}.} % 中间页眉
}
\usepackage[table,dvipsnames]{xcolor}

\usepackage{enumitem}
\setlist{nosep}

\usepackage{epsfig} % 用于插入EPS格式的图片
\usepackage{graphicx} % 提供插入图片支持
\usepackage[caption=false,font=footnotesize,labelfont=sf,textfont=sf]{subfig} % \subfloat

\usepackage{booktabs} % for \toprule, \midrule etc macros
\usepackage{colortbl}
\usepackage{multirow}
\usepackage{array}
\usepackage[hypcap=false]{caption}

\usepackage{amsmath,amssymb,bm}
\usepackage{mathpazo}
\usepackage{pifont}
\usepackage{xspace}

\usepackage{algorithmic}
\usepackage[linesnumbered,ruled]{algorithm2e}

\usepackage{url} % 使用\url命令，用于格式化和显示 URL，它会自动断行以适应页面宽度，并且通常会转义特殊字符使其能够正确显示为网址的一部分。
\usepackage[
    pagebackref=true,
    breaklinks=true,
    bookmarks=true,
    colorlinks,          % 开启彩色链接（默认状态下链接为彩色方框）
    linkcolor=magenta,       % 设置普通内联链接颜色为红色
    citecolor=magenta,      % 设置引用链接颜色为蓝色
    filecolor=cyan,   % 设置文件链接颜色为洋红色
    urlcolor=cyan,       % 设置 URL 链接颜色为青色
]{hyperref} % 提供引用能力，同时提供的\href不仅可以生成美观的可点击链接，而且可以自定义显示在文档中的文本内容，不必完全显示完整的 URL。但是不会自动断行。
\usepackage[square,comma,numbers,sort]{natbib} % 优化引用的显示方式。相比biblatex宏包兼容性更好
\usepackage{cleveref} % 优化文档内引用的使用方式

\definecolor{tabtitle}{gray}{.8}
\definecolor{ours}{gray}{.95}
\definecolor{ggray}{RGB}{127,127,127}
\definecolor{mygray}{RGB}{170,170,170}
\definecolor{mylred}{RGB}{250,130,130}
\definecolor{mylgreen}{RGB}{130,250,130}
\definecolor{mylblue}{RGB}{130,130,250}
\definecolor{reda}{RGB}{202,0,0}
\definecolor{redb}{RGB}{217,148,143}
\definecolor{myyellow}{RGB}{190,144,0}
\definecolor{mygreen}{RGB}{0,136,51}
\definecolor{myblue}{RGB}{0,102,204}

\newcolumntype{B}{!{\vrule width 1pt}}

\newcommand{\no}{\text{\ding{55}}}  %
\newcommand{\none}{—}

\newcommand{\best}[1]{\textcolor{reda}{\textbf{#1}}}
\newcommand{\second}[1]{\textcolor{mygreen}{\textbf{#1}}}
\newcommand{\third}[1]{\textcolor{myblue}{\textbf{#1}}}

\newcommand{\myModel}{{ZoomNeXt}\xspace}
\newcommand{\mymodel}{{unified collaborative pyramid network}\xspace}

\newcommand{\myrouting}{{difference-aware adaptive routing mechanism}\xspace}

\newcommand{\myMhsiu}{{Multi-head scale integration unit}\xspace}

\newcommand{\mymhsiu}{{multi-head scale integration unit}\xspace}

\newcommand{\myMHSIU}{{MHSIU}\xspace}
\newcommand{\myMHSIUs}{{MHSIUs}\xspace}

\newcommand{\myRgpu}{{Rich granularity perception unit}\xspace}

\newcommand{\myrgpu}{{rich granularity perception unit}\xspace}

\newcommand{\myRGPU}{{RGPU}\xspace}
\newcommand{\myRGPUs}{{RGPUs}\xspace}

\newcommand{\myUal}{{Uncertainty awareness loss}\xspace}
\newcommand{\myual}{{uncertainty awareness loss}\xspace}
\newcommand{\myUAL}{{UAL}\xspace}
\newcommand{\myUaL}{{Uncertainty Awareness Loss}}

\newcommand{\eg}{{\textit{e.g.},}\xspace}
\newcommand{\ie}{{\textit{i.e.},}\xspace}
\newcommand{\real}{\mathbb{R}}
\newcommand{\parhead}[1]{\noindent\textbf{#1}}

\newenvironment{authorfoot}
  {}
  {}
\newcommand{\keywords}[1]{\noindent\textbf{Keywords}~#1}

\begin{document}

\title{\myModel: A Unified Collaborative Pyramid Network for Camouflaged Object Detection}

\author{
Youwei Pang\textsuperscript{1}\protect\footnotemark[1],
Xiaoqi Zhao\textsuperscript{1}\protect\footnotemark[1],
Tian-Zhu Xiang\textsuperscript{2}\protect\footnotemark[1],\\
Lihe Zhang,~\textit{Member,~IEEE}\textsuperscript{1}\protect\footnotemark[2],
and Huchuan Lu,~\textit{Fellow,~IEEE}\textsuperscript{1}\\
{\normalsize
\textsuperscript{1}Dalian University of Technology, China,
\textsuperscript{2}Inception Institute of Artificial Intelligence and G42, UAE}\\
{\normalsize\texttt{\url{https://github.com/lartpang/ZoomNeXt}}}
}

\date{\vspace{-5ex}} % 移除日期信息

\maketitle

\thispagestyle{myheaderfooter} % 应用自定义的页眉页脚样式

\begin{authorfoot}
  \footnotetext[1]{These authors contributed equally to this work.}
  \footnotetext[2]{Corresponding author.}
\end{authorfoot}

\begin{abstract}
  Recent camouflaged object detection (COD) attempts to segment objects visually blended into their surroundings, which is extremely complex and difficult in real-world scenarios.
  Apart from the high intrinsic similarity between camouflaged objects and their background, objects are usually diverse in scale, fuzzy in appearance, and even severely occluded.
  To this end, we propose an effective unified collaborative pyramid network that mimics human behavior when observing vague images and videos, \ie zooming in and out.
  Specifically, our approach employs the zooming strategy to learn discriminative mixed-scale semantics by the multi-head scale integration and rich granularity perception units, which are designed to fully explore imperceptible clues between candidate objects and background surroundings.
  The former's intrinsic multi-head aggregation provides more diverse visual patterns.
  The latter's routing mechanism can effectively propagate inter-frame differences in spatiotemporal scenarios and be adaptively deactivated and output all-zero results for static representations.
  They provide a solid foundation for realizing a unified architecture for static and dynamic COD.
  Moreover, considering the uncertainty and ambiguity derived from indistinguishable textures, we construct a simple yet effective regularization, uncertainty awareness loss, to encourage predictions with higher confidence in candidate regions.
  Our highly task-friendly framework consistently outperforms existing state-of-the-art methods in image and video COD benchmarks.

\end{abstract}

\section{Introduction}
\label{sec:introduction}

Camouflaged objects are often ``seamlessly'' integrated into the environment by changing their appearance, coloration or pattern to avoid detection, such as chameleons, cuttlefishes and flatfishes.
This natural defense mechanism has evolved in response to their harsh living environments.
Broadly speaking, camouflaged objects also refer to objects that are extremely small in size, highly similar to the background, or heavily obscured.
They subtly hide themselves in the surroundings, making them difficult to be found, \eg soldiers wearing camouflaged uniforms and lions hiding in the grass.
Therefore, camouflaged object detection (COD) presents a significantly more intricate challenge compared to traditional salient object detection (SOD) or other object segmentation.
Recently, it has piqued ever-growing research interest from the computer vision community and facilitates many valuable real-life applications, such as
search and rescue~\cite{SINetV2},
species discovery~\cite{EarlyWEvolutionandEcology},
medical analysis (\eg
polyp segmentation~\cite{VideoPolypSegmentation, PraNet, MSNet},
lung infection segmentation~\cite{Inf-Net},
and cell segmentation~\cite{li2022tri}),
agricultural management~\cite{CODRelatedWork-PestDetection, CODRelatedWork-FruitRipenessClassification},
and defect detection~\cite{CamouflageObject-DefectIdentification, DeepCSU}.

\begin{figure}[t]
  \centering
  \includegraphics[width=\linewidth]{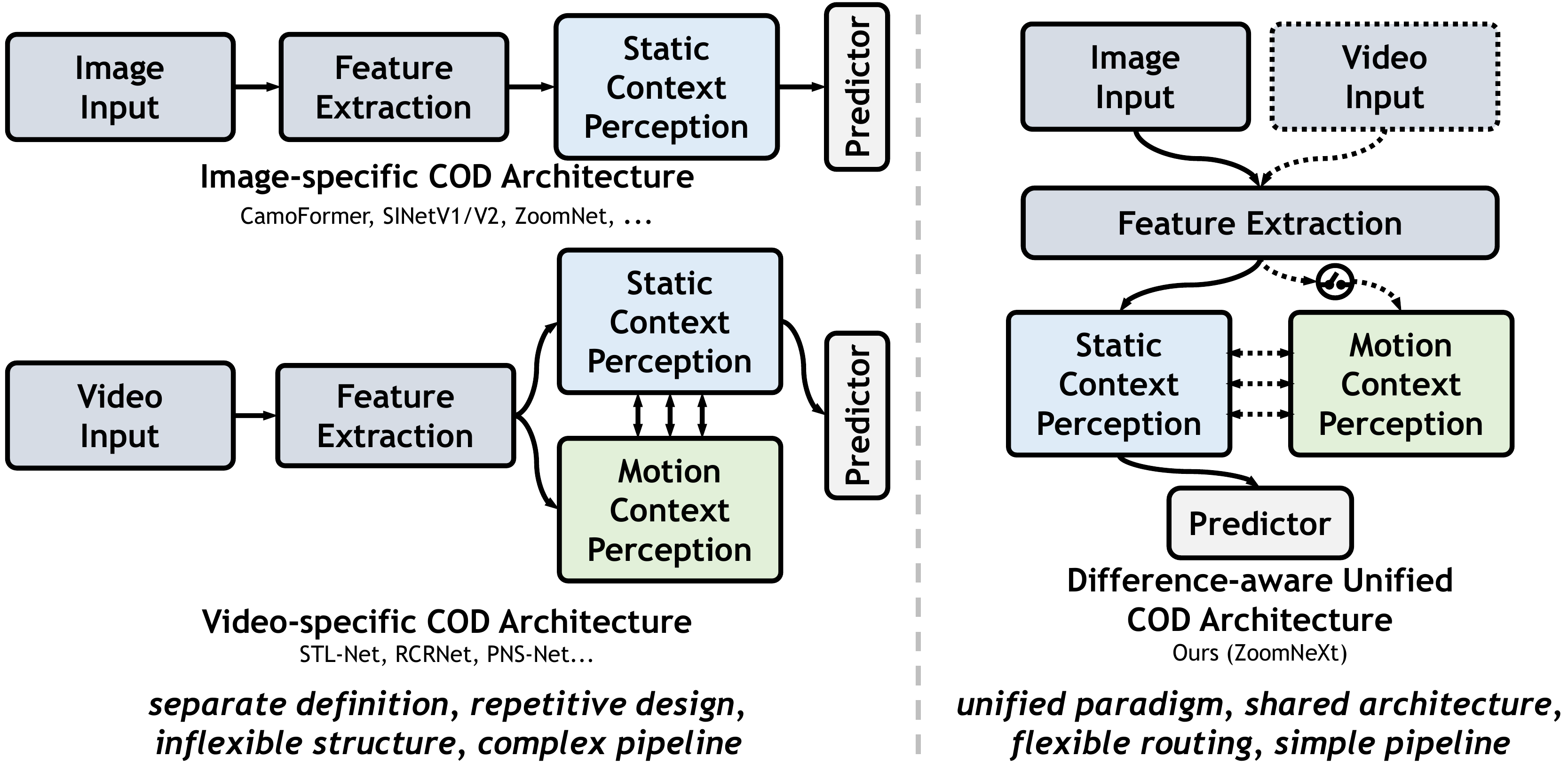}
  \caption{
    Comparison of the proposed framework with existing paradigms.
    Existing methods feed the image and video inputs into task-specific architecture to yield camouflaged object map.
    Unlike them, our \myModel, which is founded on the flexible difference-aware routing mechanism \includegraphics[height=2ex]{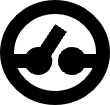},
    unifies and simplifies the processing pipeline of image and video COD tasks, which shares powerful static components without introducing redundant computation.
  }
  \label{fig:teaser}
  \vspace{-1em}
\end{figure}

Recently, numerous deep learning-based methods have been proposed and achieved significant progress.
Nevertheless, they still struggle to accurately and reliably detect camouflaged objects, due to the visual insignificance of camouflaged objects, and high diversity in scale, appearance, and occlusion.
By observing our experiments, it is found that the current COD detectors are susceptible to distractors from background surroundings.
Thus it is difficult to excavate discriminative and subtle semantic cues for camouflaged objects, resulting in the inability to clearly segment the camouflaged objects from the chaotic background and the predictions of some uncertain (low-confidence) regions.
In addition, recent works~\cite{VCOD-MoCA, VCOD-CAD, VCOD-MoCA-Mask} have attempted to introduce temporal motion cues of objects to further expand the information dimension of the COD task.
However, as shown in Fig.~\ref{fig:teaser}, the existing best solutions for image~\cite{COD-CamoFormer, SINetV2, COD-ZoomNet} and video~\cite{VCOD-MoCA-Mask} COD tasks have different information emphases (\ie the appearance information of static images and the motion information of videos), resulting in incompatibility in architecture design and feature pipeline between the two types of tasks.
The task-specific development and evolution bring repetitive exploration, inflexible cross-task structures, redundant parameters, and resource overhead.
The unification of the two tasks has a positive value in fully exploiting the inter-task commonalities and leveraging powerful shared components (\eg feature extraction and static context perception as shown in Fig.~\ref{fig:teaser}) to integrate key static cues from the image and video inputs.
In addition, it is poorly performed to apply image models to the video field because of the absence of motion perception.
And, video models based on the predefined motion information need to specifically compensate for missing auxiliary inputs and bring unnecessary computation when handling static images.
Taking these into mind, in this paper, we summarize the COD issue into three aspects:
1) \textit{How to accurately locate camouflaged objects under conditions of inconspicuous appearance and various scales?}
2) \textit{How to design a unified framework that is compatible with both image and video feature processing pipelines?}
3) \textit{How to suppress the obvious interference from the background and infer camouflaged objects more reliably?}

Intuitively, to accurately find the vague or camouflaged objects in the scene, humans may try to refer to and compare the changes in the shape or appearance at different scales by zooming in and out the image or video.
This specific behavior pattern of human beings motivates us to identify camouflaged objects by mimicking the zooming in and out strategy.
With this inspiration, in this paper, we propose a \mymodel, named \textit{ZoomNeXt}, which significantly improves the existing camouflaged object detection performance for the image and video COD tasks.
\textbf{Firstly}, for accurate object location, we employ scale space theory~\cite{ScaleSpaceTheory1,ScaleSpaceTheory2,ScaleSpaceFiltering} to imitate zooming in and out strategy.
Specifically, we design two key modules, \ie the \mymhsiu (\myMHSIU) and the \myrgpu (\myRGPU).
For the camouflaged object of interest, our model extracts discriminative features at different ``zoom'' scales using the triplet architecture, then adopts \myMHSIUs to screen and aggregate scale-specific features, and utilizes \myRGPUs to further reorganize and enhance mixed-scale features.
\textbf{Secondly}, the \myrouting of the module \myRGPU can also make full use of the inter-frame motion information to capture camouflaged objects in the video.
This flexible data-aware architecture design seamlessly unifies the processing pipeline for static images and video clips into a single convolutional model.
Thus, our model is able to mine the accurate and subtle semantic clues between objects and backgrounds under mixed scales and diverse scenes, and produce accurate predictions.
Besides, the shared weight strategy used in the proposed method achieves a good balance of efficiency and effectiveness.
\textbf{Thirdly}, the last aspect is related to reliable prediction in complex scenarios.
Although the object is accurately located, the indistinguishable texture and background will easily bring negative effects to the model learning, \eg predicting uncertain/ambiguous regions, which greatly reduces the detection performance and cannot be ignored.
This can be seen in Fig.~\ref{fig:visualation} (Row 3 and 4) and Fig.~\ref{fig:featvisualablation}.
To this end, we design a \myual (\myUAL) to guide the model training, which is only based on the prior knowledge that a good COD prediction should have a clear polarization trend.
Its characteristic of not relying on ground truth (GT) data makes it suitable for enhancing the GT-based BCE loss.
This targeted enhancement strategy can force the network to optimize the prediction of the uncertain regions during the training process, enabling our \myModel to distinguish the uncertain regions and segment the camouflaged objects reliably.

Our contributions can be summarized as follows:
\begin{enumerate}
  \item For the COD task, we propose \textit{\myModel}, which can credibly capture the objects in complex scenes by characterizing and unifying the scale-specific appearance features at different ``zoom'' scales and the purposeful optimization strategy.
  \item To obtain the discriminative feature representation of camouflaged objects, we design \myMHSIUs and \myRGPUs to distill, aggregate, and strengthen the scale-specific and subtle semantic representation for accurate COD.
  \item Through a carefully-designed \myrouting, our unified conditional architecture seamlessly combines the image and video feature pipelines, enhancing scalability and flexibility.
  \item We propose a simple yet effective optimization enhancement strategy, \myUAL, which can significantly suppress the uncertainty and interference from the background without increasing extra parameters.
  \item Without bells and whistles, our model greatly surpasses the recent 30 state-of-the-art methods on four image and two video COD data benchmarks, respectively.
\end{enumerate}

A preliminary version of this work was published in~\cite{COD-ZoomNet}.
In this paper, we extend our conference version in the following aspects.
First, we \textbf{improve image-specific \textit{ZoomNet} to image-video unified \textit{ZoomNeXt}.}
We extend the original static image COD framework to enable further compatibility with the new auxiliary video-related temporal cues.
Second, we \textbf{explore the video COD task.}
Based on the improved framework, we extend the targeted task from the single image COD task to image and video COD tasks.
Third, we \textbf{improve the performance of our method by introducing more structural extensions}, \ie~\mymhsiu and \myrgpu.
They make our approach superior not only to the conference version, but even to several recent cutting-edge competitors.
Fourth, we also \textbf{introduce and discuss more recently published algorithms for image and video tasks} so as to track recent developments in the field and make fair and reasonable comparisons.
Last but not least, we \textbf{provide more analytical experiments and visualization} to further study the effectiveness and show the characteristics of the provided components in our method including the \myMHSIU, \myRGPU, and GT-free loss function \myUAL, which helps to better understand their behavior.

\section{Related Work}

\subsection{Camouflaged Object Detection (COD)}

Before the visual detection of camouflaged objects as objects of interest becomes an independent task in computer vision, research on such objects has a long history in biology~\cite{CamouflageObject-AnimalCamouflage, CamouflageObject-Perception, CamouflageObject-HowWork}.
This behavior of creatures in nature can be regarded as the result of natural selection and adaptation.
In fact, in human life and other parts of society, it also has a profound impact, \eg arts, popular culture, and design.
In the computer vision field, research on camouflaged objects is often associated with visual salience analysis~\cite{CamouflageObject-VisualSalience, CamouflageObject-SalinecyMap, CamouflageObject-Breaking, CamouflageObject-FeaturesFusion} based on the mechanism of human visual attention.

Different from the established salient object detection (SOD)~\cite{CMMSODSurvey, WWGSODSurvey, RGBDSODSurveyZhou, DeepLightField} task for the general observation paradigm (\ie finding visually prominent objects), the COD task pays more attention to the undetectable objects (mainly because of too small size, occlusion, concealment or self-disguise).
Due to the differences in the attributes of the objects of interest, the goals of the two tasks are different.
The difficulty and complexity of the COD task far exceed the SOD task due to the high similarity between the object and the environment~\cite{ChenGXL23, DongPGXWX23, ZhangCoCOD2023}.
Therefore, it is necessary to establish models based on the essential requirements and specific data of the task to learn the special knowledge.
Although it has practical significance in both the image and video fields, the COD task is still a relatively under-explored computer vision problem.
Some valuable attempts have been made in recent years.
The pioneering works~\cite{CAMO,COD10K,SLSR} collect and release several important image COD datasets CAMO~\cite{CAMO}, COD10K~\cite{COD10K}, and NC4K~\cite{SLSR}.
And, another vanguard~\cite{VCOD-MoCA-Mask} extends the task to the video COD task and introduces a large-scale comprehensive dataset MoCA-Mask~\cite{VCOD-MoCA-Mask}.
These datasets have become representative data benchmarks for the COD task nowadays and built a solid foundation for a large number of subsequent studies.

Recent works~\cite{CAMO, COD-MGL, SLSR, COD-BSA-Net, COD-BGNet, COD-FEDER, COD-DGNet} construct the multi-task learning framework in the prediction process of camouflaged objects and introduce some auxiliary tasks like classification, edge detection, and object gradient estimation.
Among them, \cite{COD-FEDER} also attempts to address the intrinsic similarity of foreground and background for COD in the frequency domain.
Some uncertainty-aware methods~\cite{UJSC, COD-UGTR, zhang2023predictive} are proposed to model and cope with the uncertainty in data annotation or COD data itself.
In the other methods~\cite{COD-PFNet, COD-C2FNet}, contextual feature learning also plays an important role.
And \cite{COD-SegMaR} introduces an explicit coarse-to-fine detection and iterative refinement strategy to refine the segmentation effect.
Compared to convolutional neural networks, the vision transformer can encode global contextual information more effectively and also change the way existing methods~\cite{COD-CamoFormer, COD-FSPNet, COD-HiTNet, COD-MSCAF-Net, COD-SARNet, VCOD-MoCA-Mask} perceive contextual information.
Specifically, \cite{COD-CamoFormer, COD-SARNet} are based on the strategy of dividing and merging foreground and background to accurately distinguish highly similar foreground and background.
\cite{VCOD-MoCA-Mask} utilizes the multi-scale correlation aggregation and the spatiotemporal transformer to construct short-term and long-term information interactions, and optimize temporal segmentation.
\cite{COD-FSPNet} aims to enhance locality modeling of the transformer-based model and the feature aggregation in decoders, while \cite{COD-HiTNet} preserves detailed clues through the high-resolution iterative feedback design.
In addition, there are also a number of bio-inspired methods, such as~\cite{COD10K, SINetV2, COD-MirrorNet}.
They capture camouflaged objects by imitating the behavior process of hunters or changing the viewpoint of the scene.

Although our method can also be attributed to the bio-inspired paradigm, ours is different from the above methods.
Our method simulates the behavior of humans to understand complex images by zooming in and out strategy.
The proposed method explores the scale-specific and imperceptible semantic features under the mixed scales for accurate predictions, with the supervision of BCE and our proposed \myUAL.
At the same time, through the adaptive perception ability of the temporal difference, the proposed method also harmonizes both image and video data in a single framework.
Accordingly, our method achieves a more comprehensive understanding of the diverse scene, and accurately and robustly segments the camouflaged objects from the complex background.

\subsection{Conditional Computation}

Conditional computation~\cite{DLRep-Bengio} refers to a series of carefully constructed algorithms in which each input sample actually uses only a subset of feature processing nodes.
It can further reduce computation, latency, or power on average, and increase model capacity without a proportional increase in computation.
As argued in~\cite{DLRep-Bengio}, \textit{sparse activations} and \textit{multiplicative connections} may be useful ingredients of conditional computation.
Over the past few years, it has become an important solution to the time-consuming and computationally expensive training and inference of deep learning models.
Specifically, as a typical paradigm, the mixture of experts (MoE) technique~\cite{MoE} based on sparse selection has shown great potential on a variety of different types of tasks, including language modeling and machine translation~\cite{SparselyGatedMoE}, multi-task learning~\cite{DSelect-k-MoE}, image classification~\cite{Sparse-MLP-MoE, V-MoE}, and vision-language model~\cite{VL-MoE}.
Existing methods mainly rely on MoE and gating strategies to realize dynamic routing of feature flow nodes, and focus on the distinction between different samples of homogeneous data.
In the proposed scheme, we propose a new form of conditional computation for differentiated processing of image and video data.
Specifically, we construct adaptive video-specific bypasses using the inter-frame difference in video, which is driven by the inter-frame motion information and only responsible for perceiving and excavating dynamic cues, and will be adaptively deactivated and output all-zero results when facing the static image.
Such a design decouples universal static perception and specific motion perception, resulting in good compatibility across image and video COD tasks.

\begin{figure*}[!t]
  \centering
  \includegraphics[width=0.9\linewidth]{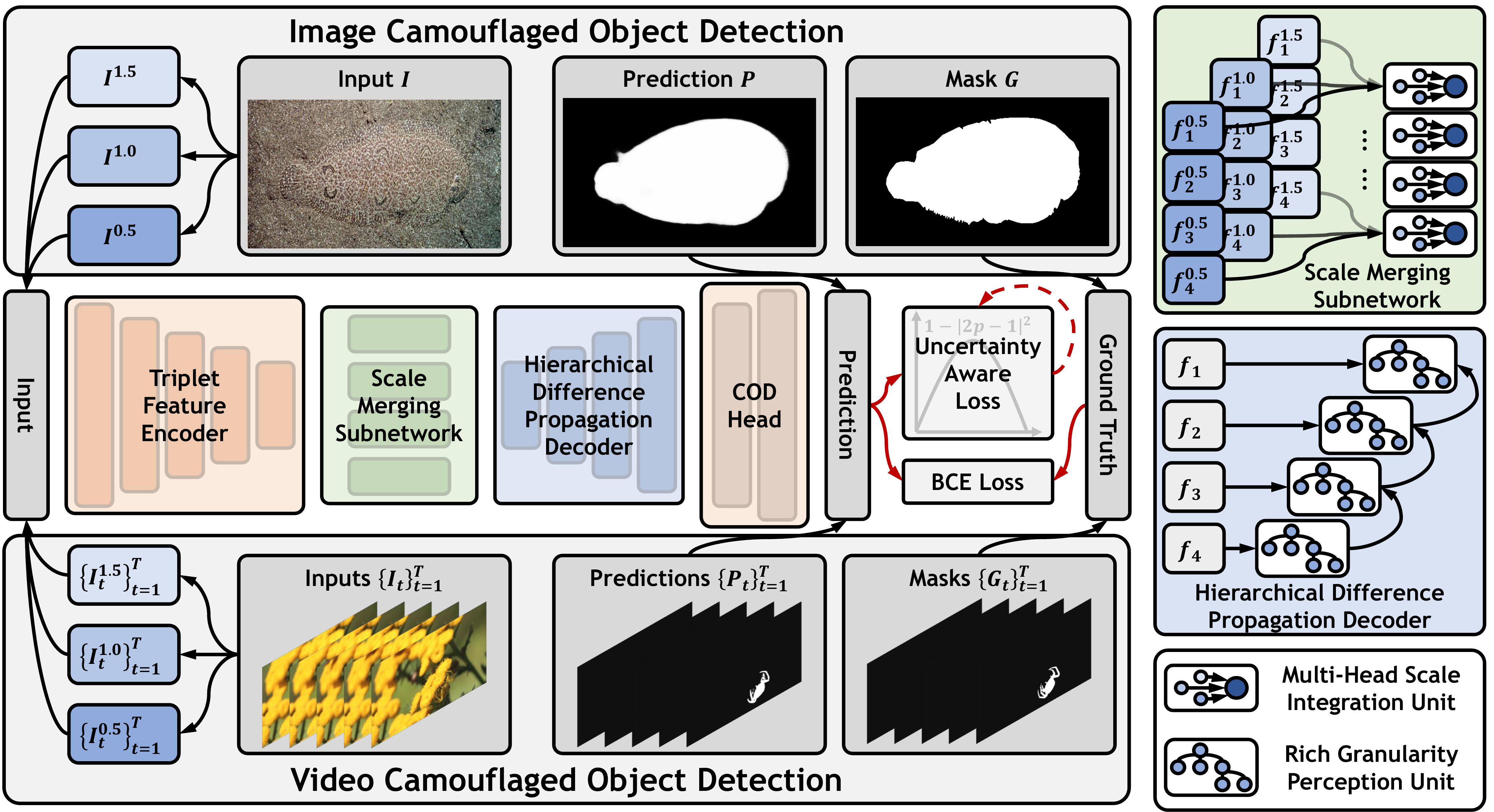}
  \caption{Overall framework of the proposed \myModel.
    The shared triplet feature encoder is used to extract multi-level features corresponding to different input ``zoom'' scales.
    At different levels of the scale merging subnetwork, \myMHSIUs are adopted to screen and aggregate the critical cues from different scales.
    Then the fused features are gradually integrated through the top-down up-sampling path in the hierarchical difference propagation decoder.
    \myRGPUs further enhance the feature discrimination by constructing a multi-path structure inside the features.
    Finally, the probability map of the camouflaged object corresponding to the input image or frame can be obtained.
    In the training stage, the binary cross entropy and the proposed \myual are used as the loss function.
  }
  \label{fig:net}
  \vspace{-1em}
\end{figure*}

\subsection{Scale Space Integration}

The scale-space theory aims to promote an optimal understanding of image structure, which is an extremely effective and theoretically sound framework for addressing naturally occurring scale variations.
Its ideas have been widely used in computer vision, including the image pyramid~\cite{ImagePyramid, MultiScaleNet} and the feature pyramid~\cite{FPN}.
Due to the structural and semantic differences at different scales, the corresponding features play different roles.
However, the commonly-used inverted pyramid-like feature extraction structures~\cite{SPP, FasterRCNN, MSAPS} often cause the feature representation to lose too much texture and appearance details, which are unfavorable for dense prediction tasks~\cite{FCN, Unet} that emphasize the integrity of regions and edges.
Thus, some recent CNN-based COD methods~\cite{COD10K, COD-C2FNet, COD-PFNet, COD-MGL} and SOD methods~\cite{PoolNet, MINet, GateNet, DMRA-TIP, HDFNet, DANet, SSLSOD, WSRSSOD-huang2022scribble, RGBDSOD-bi2023cross} explore the combination strategy of inter-layer features to enhance the feature representation.
These bring some positive gains for accurate localization and segmentation of objects.
However, for the COD task, the existing approaches overlook the performance bottleneck caused by the ambiguity of the structural information of the data itself which makes it difficult to be fully perceived at a single scale.
Different from them, we mimic the zoom strategy to synchronously consider differentiated relationships between objects and background at multiple scales, thereby fully perceiving the camouflaged objects and confusing scenes.
Besides, we also further explore the fine-grained feature scale space between channels.

\section{Method}

In this section, we first elaborate on the overall architecture of the proposed method, and then present the details of each module and the \myual function.

\subsection{Overall Architecture}\label{sec:architecture}

\parhead{Preliminaries}
Let $\mathbf{I} \in \real^{3 \times H \times W}$ and $\{\mathbf{I}_t \in \real^{3 \times H \times W}\}^{T}_{t=1}$ be the input static image and the input video clip with $T$ frames of the network, where $3$ is the number of color channels and $H, W$ are the height and width.
Our network is to generate a gray-scale map $\mathbf{P}$ or clip $\{\mathbf{P}_t\}^{T}_{t=1}$ with values ranging from 0 to 1, which reflects the probability that each location may be part of the camouflaged object.

\parhead{Zooming Strategy}
The overall architecture of the proposed method is illustrated in Fig.~\ref{fig:net}.
Inspired by the zooming strategy from human beings when observing confusing scenes, we argue that different zoom scales often contain their specific information.
Aggregating the differentiated information on different scales will benefit exploring the inconspicuous yet valuable clues from confusing scenarios, thus facilitating COD.
To implement it, intuitively, we resort to the image pyramid.
Specifically, we customize an image pyramid based on the single scale input to identify the camouflaged objects.
The scales are divided into a main scale (\ie the input scale) and two auxiliary scales.
The latter is obtained by re-scaling the input to imitate the operation of zooming in and out.

\parhead{Feature Processing}
We utilize the shared triplet feature encoder to extract features on different scales and feed them to the scale merging subnetwork.
To integrate these features that contain rich scale-specific information, we design a series of multi-head scale integration units (MHSIUs) based on the attention-aware filtering mechanism.
Thus, these auxiliary scales are integrated into the main scale, \ie information aggregation of the ``zoom in and out'' operation.
This will largely enhance the model to distill critical and informative semantic cues for capturing difficult-to-detect camouflaged objects.
After that, we construct rich granularity perception units (\myRGPUs) to gradually integrate multi-level features in a top-down manner to enhance the mixed-scale feature representation.
It further increases the receptive field range and diversifies feature representation within the module.
The captured fine-grained and mixed-scale clues promote the model to accurately segment the camouflaged objects in the chaotic scenes.

\parhead{Loss Improvement}
To overcome the uncertainty in the prediction caused by the inherent complexity of the data, we design an \myual (\myUAL) to assist the BCE loss, enabling the model to distinguish these uncertain regions and produce an accurate and reliable prediction.

\subsection{Triplet Feature Encoder}\label{sec:encoder}

We start by extracting deep features through a shared triplet feature encoder for the group-wise inputs, which consists of a feature extraction network and a channel compression network.
The feature extraction network is constituted by the commonly-used ResNet~\cite{Resnet}, EfficientNet~\cite{EfficientNet}, or PVTv2~\cite{PVTv2} that removes the classification head.
And the channel compression network is cascaded to further optimize computation and obtain a more compact feature.
For the trade-off between efficiency and effectiveness, the main scale and the two auxiliary scales are empirically set to $1.0\times$, $1.5\times$, and $0.5\times$.
And, three sets of 64-channel feature maps corresponding to three input scales are produced by these structures, \ie $\{f^{k}_{i}\}_{i=1}^{5}, \, k \in \{0.5, 1.0, 1.5\}$.
Next, these features are fed successively to the multi-head scale merging subnetwork and the hierarchical difference propagation decoder for subsequent processing.

\subsection{Scale Merging Subnetwork}
\label{sec:scale_merging}

\begin{figure}[t]
  \centering
  \includegraphics[width=0.9\linewidth]{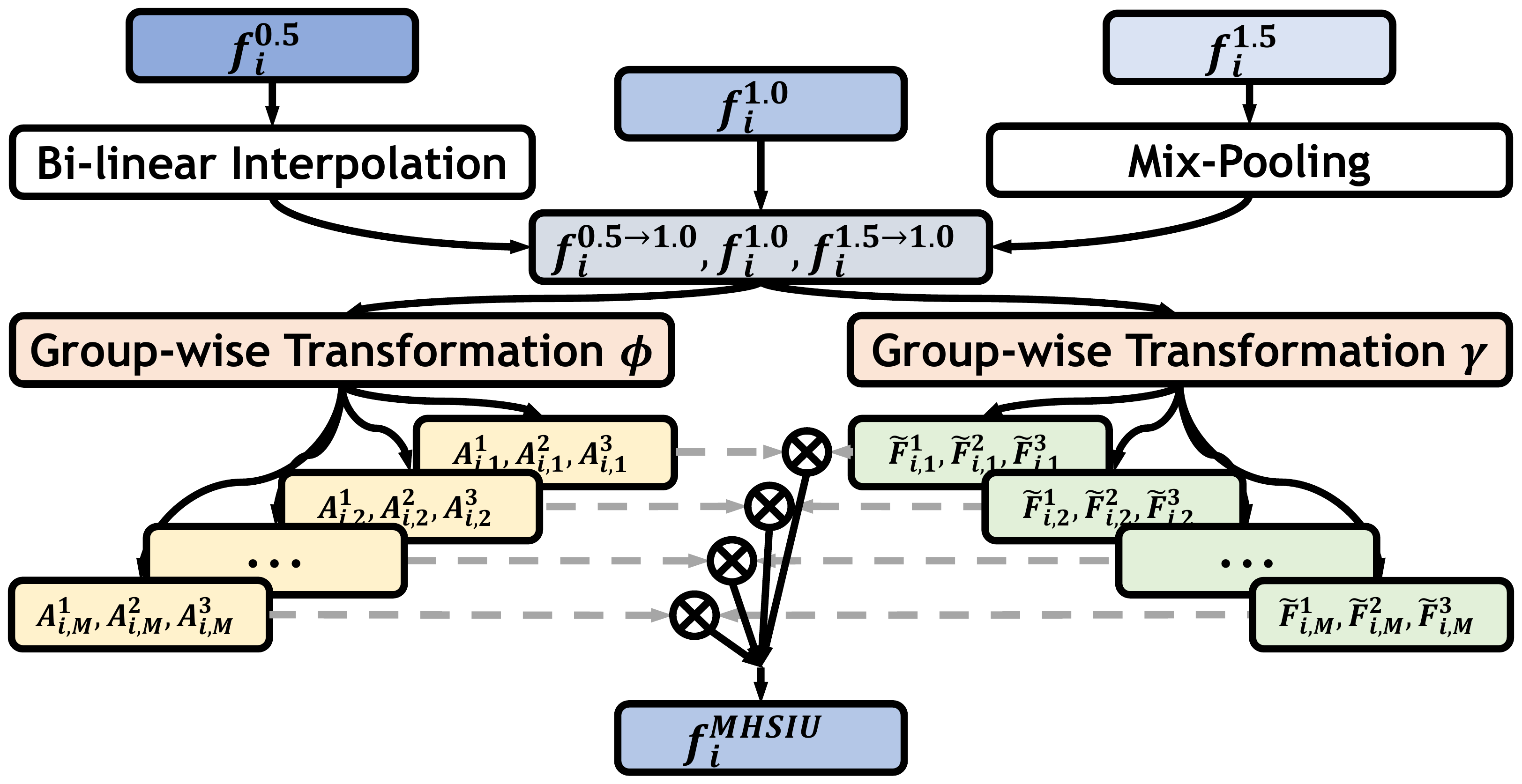}
  \caption{Illustration of the \mymhsiu.
    $\otimes$ is the element-wise multiplication.
    $\phi$ and $\gamma$ are the parameters of the two separated group-wise transformation layers.
    More details can be found in Sec.~\ref{sec:scale_merging}.}
  \label{fig:mhsiu}
  \vspace{-1em}
\end{figure}

We design an attention-based \mymhsiu (MHSIU) to screen (weight) and combine scale-specific information, as shown in Fig.~\ref{fig:mhsiu}.
Several such units make up the scale merging subnetwork.
Through filtering and aggregation, the expression of different scales is self-adaptively highlighted.
This component is described in detail next.

\parhead{Scale Alignment}
Before scale integration, the features $f^{1.5}_{i}$ and $f^{0.5}_{i}$ are first resized to be consistent resolution with the main scale feature $f^{1.0}_{i}$.
Specifically, for $f^{1.5}_{i}$, we use a hybrid structure of ``max-pooling + average-pooling'' inspired by~\cite{CBAM} to down-sample it, which helps to preserve the effective and diverse responses for camouflaged objects in high-resolution features.
For $f^{0.5}_{i}$, we directly up-sample it by the bi-linear interpolation.
Then, these features are fed into the subsequent transformation layers.

\parhead{Multi-Head Spatial Interaction}
Unlike the form of spatial attention in our conference version~\cite{COD-ZoomNet} that relies heavily on a single pattern, here we perform parallel independent transformations on $M$ groups of feature maps, which is inspired by the structure of the multi-head paradigm in Transformer~\cite{Transformer}.
This design helps extend the model's ability to mine multiple fine-grained spatial attention patterns in parallel and diversify the representation of the feature space.
Specifically, several three-channel feature maps are calculated through a series of convolutional layers.
After the cascaded $\texttt{softmax}$ activation layer in each attention group, the attention map $A^k_m$ corresponding to each scale can be obtained from these groups and used as respective weights for the final integration, where scale group index $k \in \{1, 2, 3\}$ and attention group index $m \in \{1, 2, \dots, M\}$.
The process is formulated as:
\begin{equation}
  \begin{split}
    F_{i} & = \left[\mathcal{U}(f^{0.5}_{i}), f^{1.0}_{i}, \mathcal{D}(f^{1.5}_{i}) \right], \\
    \hat{F}_{i} & = \{ \texttt{trans}( F_{i}, \phi^{m} ) \}_{m=1}^{M}, \\
    A_{i} & = \{ \texttt{softmax}(\hat{F}_{i,m}) \}_{m=1}^{M}, \\
    \tilde{F}_{i} & = \{ \texttt{trans}( F_{i}, \gamma^{m} ) \}_{m=1}^{M}, \\
    f^{MHSIU}_{i} & = \{ A^{1}_{i,m} \otimes \tilde{F}^{1}_{i,m}
    + A^{2}_{i,m} \otimes \tilde{F}^{2}_{i,m}
    + A^{3}_{i,m} \otimes \tilde{F}^{3}_{i,m}
    \}_{m=1}^{M},
  \end{split}
  \label{equ:fusion}
\end{equation}
where $\left[ \star \right]$ represents the concatenation operation.
$\mathcal{U}$ and $\mathcal{D}$ refer to the bi-linear interpolation and hybrid adaptive pooling operations mentioned above, respectively.
$\texttt{trans}(\star, \phi)$ and $\texttt{trans}(\star, \gamma)$ indicate the several simple linear transformation layers with the parameters $\phi$ and $\gamma$ in the attention generator.
And $\otimes$ is the element-wise multiplication operation.
Note that some linear, normalization, and activation layers are not shown in Equ.~\ref{equ:fusion} and Fig.~\ref{fig:mhsiu} for simplicity.
And features enhanced in different groups are concatenated along the channel dimension and fed into a decoder for further processing.
These designs aim to adaptively and selectively aggregate the scale-specific information to explore subtle but critical semantic cues at different scales, boosting the feature representation.

\begin{figure}[t]
  \centering
  \includegraphics[width=0.9\linewidth]{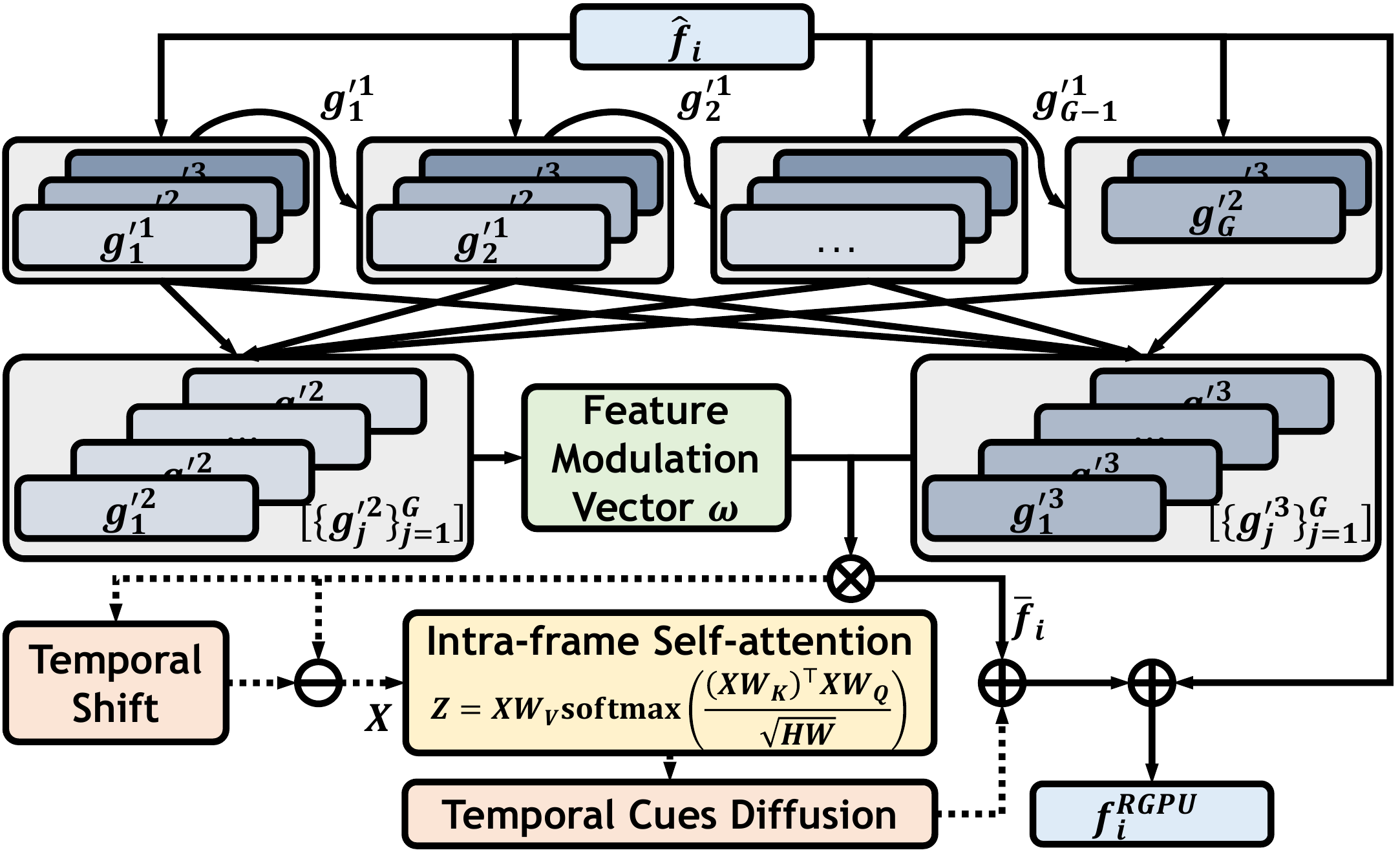}
  \caption{\myRgpu
    where $\otimes$, $\oplus$, and $\ominus$ are the element-wise multiplication, addition, and subtraction.
    Group-wise interaction and channel-wise modulation are used to explore the discriminative and valuable semantics from different channels.
    Each feature group is executed sequentially and the latter one integrates part of the features of the previous one before the feature transformation.
    The temporal shifting operation shifts the frame feature maps along the temporal dimension and some temporal convolutional layers diffuse the temporal cues as stated in Sec.~\ref{sec:decoder}.
  }
  \label{fig:decoder}
  \vspace{-1em}
\end{figure}

\subsection{Hierarchical Difference Propagation Decoder}
\label{sec:decoder}

After MHSIUs, the auxiliary-scale information is integrated into the main-scale branch.
Similar to the multi-scale case, different channels also contain differentiated semantics.
Thus, it is necessary to excavate valuable clues contained in different channels.
To this end, we design the \myrgpu (\myRGPU) to conduct information interaction and feature refinement between channels, which strengthen features from coarse-grained group-wise iteration to fine-grained channel-wise modulation in the decoder, as depicted in Fig.~\ref{fig:decoder}.

\parhead{Input}
The input $\hat{f}_{i}$ of the RGPU$_i$ contains the multi-scale fused feature $f^{MHSIU}_{i}$ from the MHSIU$_i$ and the feature $f^{RGPU}_{i+1}$ from the RGPU$_{i+1}$ as $\hat{f}_{i} = f^{MHSIU}_{i} + \mathcal{U}(f^{RGPU}_{i+1})$.

\parhead{Group-wise Iteration}
We adopt $1 \times 1$ convolution to extend the channel number of feature map $\hat{f}_{i}$.
The features are then divided into $G$ groups $\{g_{j}\}_{j=1}^{G}$ along the channel dimension.
Feature interaction between groups is carried out iteratively.
Specifically, the first group $g_{1}$ is split into three feature sets $\{{g'}_{1}^{k}\}_{k=1}^{3}$ after a convolution block.
Among them, the ${g'}_{1}^{1}$ is adopted for information exchange with the next group, and the other two are used for channel-wise modulation.
In the $j^{th}$ ($1<j<G$) group, the feature $g_{j}$ is concatenated with the feature ${g'}_{j-1}^{1}$ from the previous group along the channel, followed by a convolution block and a split operation, which similarly divides this feature group into three feature sets.
It is noted that the output of the group $G$ with a similar input form to the previous groups only contains ${g'}_{G}^{2}$ and ${g'}_{G}^{3}$.
Such an iterative mixing strategy strives to learn the critical clues from different channels and obtain a powerful feature representation.
The iteration structure of feature groups in \myRGPU is actually equivalent to an integrated multi-path kernel pyramid structure with partial parameter sharing.
Such a design further enriches the ability to explore diverse visual cues, thereby helping to better perceive objects and refine predictions. The iteration processing is also listed in Alg.~\ref{alg:iterationinhmu}.

\begin{algorithm}[t]
  \footnotesize
  \caption{The iteration process in the \myRGPU.}
  \label{alg:iterationinhmu}

  \KwIn{
  $\{g_j\}_{j=1}^{G}$: feature groups;
  $G \geq 2$: the number of groups;
  $\texttt{CBR}_{C_o \times C_i}$: ``$3 \times 3$\texttt{Conv-BN-ReLU}'' units with input and output channel numbers of $C_i$ and $C_o$;
  $\texttt{split}$ and $\texttt{concat}$: channel-wise splitting and concatenating;
  }
  \KwOut{
  $\{{g'}^2_j\}_{j=1}^{G}$: the feature set for generating the modulation vector $\omega$;
  $\{{g'}^3_j\}_{j=1}^{G}$: the feature set used to be modulated and generate the final output of the \myRGPU;
  }
  \For{$i \gets 1, G$}{
  \uIf(
  \tcc*[f]{Group $1$}
  ){
  $i = 1$
  }{
  $f \gets \texttt{CBR}^{i}_{3C \times C}(g_i)$\;
  ${g'}^1_{i}, {g'}^2_{i}, {g'}^3_{i} \gets \texttt{split}(f)$\;
  ${g'}^1_{prev} \gets {g'}^1_{i}$\;
  }
  \uElseIf(
  \tcc*[f]{Group $G$}
  ){
  $i = G$
  }{
  $f \gets \texttt{CBR}^{i}_{2C \times 2C}(\texttt{concat}(g_i, {g'}^1_{prev}))$\;
  ${g'}^2_{i}, {g'}^3_{i} \gets \texttt{split}(f)$\;
  }
  \Else(
  \tcc*[f]{Group $i$, $1 < i < G$}
  ){
  $f \gets \texttt{CBR}^{i}_{3C \times 2C}(\texttt{concat}(g_i, {g'}^1_{prev}))$\;
  ${g'}^1_{i}, {g'}^2_{i}, {g'}^3_{i} \gets \texttt{split}(f)$\;
  ${g'}^1_{prev} \gets {g'}^1_{i}$\;
  }
  }
\end{algorithm}

\parhead{Channel-wise Modulation}
The features $[\{{g'}_{j}^{2}\}_{j=1}^{G}]$ are concatenated and converted into the feature modulation vector $\mathbf{\omega}$ by a small convolutional network, which is employed to weight another concatenated feature as $\bar{f}_{i} = \omega \cdot [\{{g'}_{j}^{3}\}_{j=1}^{G}]$.

\parhead{Difference-based Conditional Computation}
Because the inter-frame difference can directly reflect the temporal motion cues of the camouflaged object in the video, we design the \myrouting to realize the video-specific inter-frame information propagation and seamlessly unify the image and video COD tasks.
Specifically, the temporal $\texttt{shift}$~\cite{TSM,GSM} operation is first applied to the features from the group-wise iteration and the channel-wise modulation.
After shifting, the feature map of the starting time step in the video clip is moved to the end, while the remaining feature maps move forward by one position.
Consequently, we derive the differential representation $X = \texttt{shift}(\bar{f}_{i}) - \bar{f}_{i}$ between adjacent frames.
To enhance the motion cues of the object of interest between frames, the intra-frame self-attention is performed as $Z = XW_{V} \texttt{softmax}(\frac{(XW_{K})^{\top} XW_{Q}}{\sqrt{HW}})$.
Furthermore, the following $T \times 3 \times 3$ convolutional layers are imposed to these temporal cues to realize fully connected diffusion of information inside the video clip and results are added to the original feature $\bar{f}_{i}$.
For the static image, this pipeline is not functional and outputs an all-zero tensor, thus maintaining the original static information flow.
The activation of this pipeline relies on a specific input condition, \textit{i.e.}, whether the input is sequence features with the temporal difference.

\begin{figure}[t]
  \centering
  \includegraphics[width=0.8\linewidth]{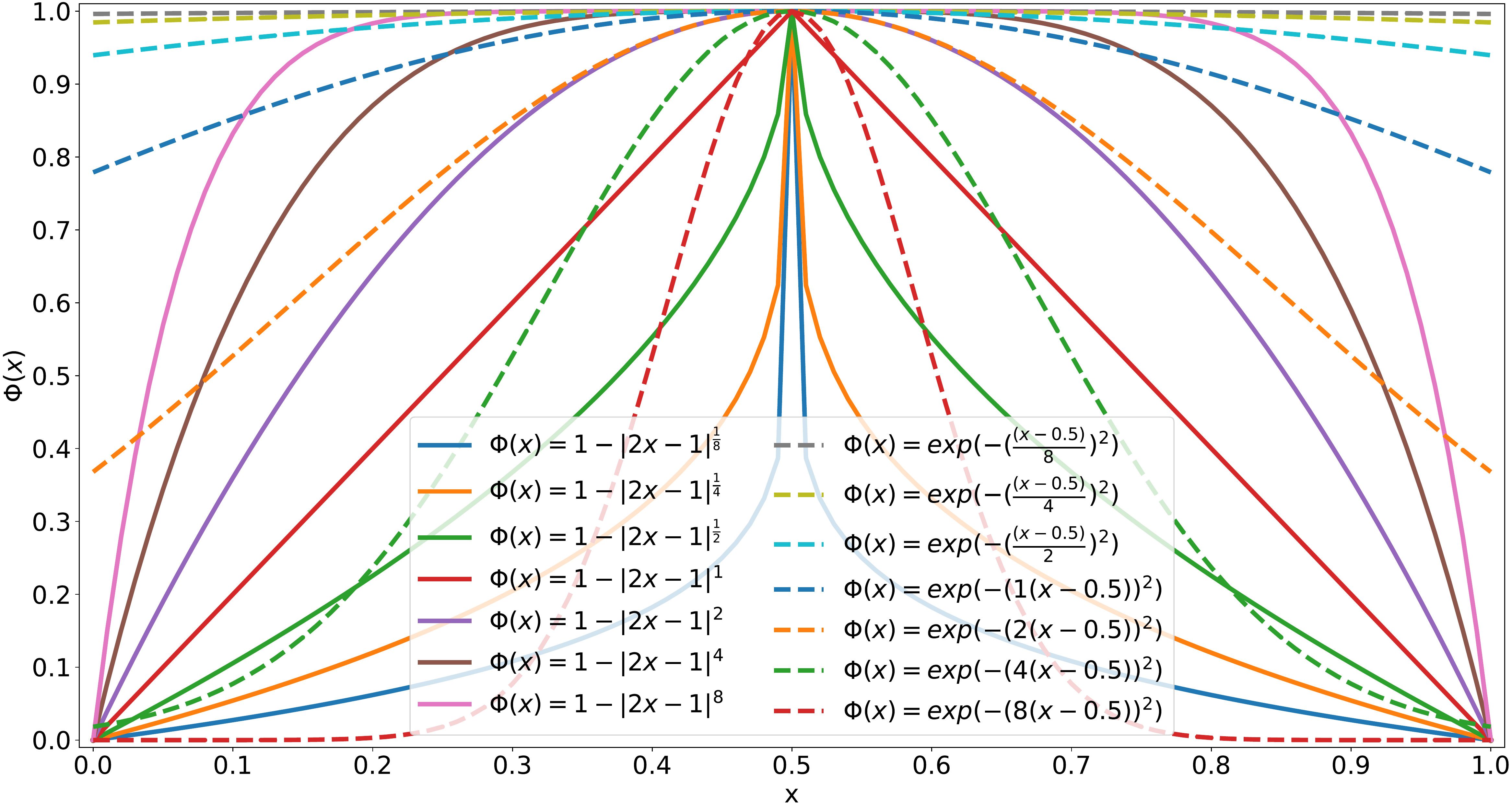}
  \caption{Curves of different forms of the proposed \myUAL.}
  \label{fig:loss_form}
  \vspace{-1em}
\end{figure}

\parhead{Output}
The output of the \myRGPU can be obtained by the stacked activation, normalization, and convolutional layers, which are defined as $f^{RGPU}_{i} = \texttt{fuse}(\hat{f}_{i} + \bar{f}_{i})$.
Based on cascaded \myRGPUs and several stacked convolutional layers, a single-channel logits map is obtained.
The final confidence map $\mathbf{P}$ or map group $\{\mathbf{P}_t\}^{T}_{t=1}$ that highlights the camouflaged objects is then generated by a \texttt{sigmoid} function.

\begin{table*}[t]
  \centering
  \caption{Comparisons of different methods based on different backbones on static image COD datasets. $^{\star}$: Using more datasets. The best three results are highlighted in {\color{reda} \textbf{red}}, {\color{mygreen} \textbf{green}} and {\color{myblue} \textbf{blue}}.}
  \resizebox{\linewidth}{!}{%
    \begin{tabular}{l|l|cc|ccccc|ccccc|ccccc|ccccc}
  \toprule[2pt]
  \rowcolor{tabtitle}
                                            &                                     &                              &                          & \multicolumn{5}{c|}{\textbf{CAMO}} & \multicolumn{5}{c|}{\textbf{CHAMELEON}} & \multicolumn{5}{c|}{\textbf{COD10K}} & \multicolumn{5}{c}{\textbf{NC4K}}                                                                                                                                                                                                                                                                                                                                                                                              \\
  \rowcolor{tabtitle}
  \multirow{-2}{*}{Model}                   & \multirow{-2}{*}{Backbone}          & \multirow{-2}{*}{Input Size} & \multirow{-2}{*}{Param.} & S$_{m}$ $\uparrow$                 & F$^{\omega}_{\beta}$ $\uparrow$         & MAE $\downarrow$                     & F$_{\beta}$ $\uparrow$            & E$_{m}$ $\uparrow$ & S$_{m}$ $\uparrow$ & F$^{\omega}_{\beta}$ $\uparrow$ & MAE $\downarrow$ & F$_{\beta}$ $\uparrow$ & E$_{m}$ $\uparrow$ & S$_{m}$ $\uparrow$ & F$^{\omega}_{\beta}$ $\uparrow$ & MAE $\downarrow$ & F$_{\beta}$ $\uparrow$ & E$_{m}$ $\uparrow$ & S$_{m}$ $\uparrow$ & F$^{\omega}_{\beta}$ $\uparrow$ & MAE $\downarrow$ & F$_{\beta}$ $\uparrow$ & E$_{m}$ $\uparrow$ \\ \midrule[1pt]
  \multicolumn{24}{c}{\textbf{Convolutional Neural Network based Methods}}                                                                                                                                                                                                                                                                                                                                                                                                                                                                                                                                                                                                                         \\ \midrule[1pt]
  SINet$^{20}$~\cite{COD10K}                & ResNet-50~\cite{Resnet}             & $352 \times 352$             & 48.947M                  & 0.745                              & 0.644                                   & 0.091                                & 0.702                             & 0.829              & 0.872              & 0.806                           & 0.034            & 0.827                  & 0.946              & 0.776              & 0.631                           & 0.043            & 0.679                  & 0.874              & 0.808              & 0.723                           & 0.058            & 0.769                  & 0.883              \\
  C$^2$FNet$^{21}$~\cite{COD-C2FNet}        & ResNet-50~\cite{Resnet}             & $352 \times 352$             & 28.411M                  & 0.796                              & 0.719                                   & 0.080                                & 0.762                             & 0.864              & 0.888              & 0.828                           & 0.032            & 0.844                  & 0.946              & 0.813              & 0.686                           & 0.036            & 0.723                  & 0.900              & 0.838              & 0.762                           & 0.049            & 0.794                  & 0.904              \\
  SINetV2$^{21}$~\cite{SINetV2}             & Res2Net-50~\cite{Res2Net}           & $352 \times 352$             & 26.976M                  & 0.820                              & 0.743                                   & 0.070                                & 0.782                             & 0.895              & 0.888              & 0.816                           & 0.030            & 0.835                  & 0.961              & 0.815              & 0.680                           & 0.037            & 0.718                  & 0.906              & 0.847              & 0.770                           & 0.048            & 0.805                  & 0.914              \\
  SegMaR$^{22}$~\cite{COD-SegMaR}           & ResNet-50~\cite{Resnet}             & $352 \times 352$             & 56.215M                  & 0.815                              & 0.753                                   & 0.071                                & 0.795                             & 0.884              & 0.906              & 0.860                           & 0.025            & 0.872                  & 0.959              & 0.833              & 0.724                           & 0.034            & 0.757                  & 0.906              & 0.841              & 0.781                           & 0.046            & 0.821                  & 0.907              \\
  CamoFormer-R$^{22}$~\cite{COD-CamoFormer} & ResNet-50~\cite{Resnet}             & $352 \times 352$             & 71.403M                  & 0.817                              & 0.752                                   & 0.067                                & 0.792                             & 0.885              & 0.898              & 0.847                           & 0.025            & 0.867                  & 0.956              & 0.838              & 0.724                           & 0.029            & 0.753                  & \third{0.930}      & 0.855              & 0.788                           & 0.042            & 0.821                  & 0.914              \\
  \rowcolor{ours}
  Ours$^{23}$                               & ResNet-50~\cite{Resnet}             & $352 \times 352$             & 28.458M                  & 0.822                              & 0.760                                   & 0.069                                & 0.797                             & 0.885              & \best{0.912}       & \third{0.863}                   & \best{0.020}     & \second{0.878}         & \best{0.969}       & 0.855              & 0.758                           & \third{0.026}    & 0.791                  & 0.926              & 0.869              & 0.808                           & 0.038            & 0.836                  & 0.925              \\
  \midrule[0.5pt]
  DGNet$^{23}$~\cite{COD-DGNet}             & EfficientNet-B4~\cite{EfficientNet} & $352 \times 352$             & 18.113M                  & \third{0.838}                      & 0.768                                   & \third{0.057}                        & 0.805                             & \third{0.914}      & 0.890              & 0.816                           & 0.029            & 0.834                  & 0.956              & 0.822              & 0.692                           & 0.033            & 0.727                  & 0.911              & 0.857              & 0.783                           & 0.042            & 0.813                  & 0.922              \\
  \rowcolor{ours}
  Ours$^{23}$                               & EfficientNet-B4~\cite{EfficientNet} & $352 \times 352$             & 21.381M                  & \second{0.859}                     & \second{0.815}                          & \second{0.049}                       & \second{0.845}                    & \second{0.920}     & \second{0.911}     & \second{0.864}                  & \best{0.020}     & \third{0.877}          & \second{0.965}     & \second{0.868}     & \second{0.789}                  & \second{0.023}   & \second{0.818}         & \second{0.937}     & \second{0.878}     & \second{0.832}                  & \second{0.035}   & \second{0.857}         & \second{0.932}     \\
  \rowcolor{ours}
  Ours$^{23}$                               & EfficientNet-B4~\cite{EfficientNet} & $384 \times 384$             & 21.381M                  & \best{0.867}                       & \best{0.824}                            & \best{0.046}                         & \best{0.852}                      & \best{0.925}       & \second{0.911}     & \best{0.865}                    & \best{0.020}     & \best{0.879}           & \third{0.964}      & \best{0.875}       & \best{0.797}                    & \best{0.021}     & \best{0.824}           & \best{0.941}       & \best{0.884}       & \best{0.837}                    & \best{0.032}     & \best{0.862}           & \best{0.939}       \\
  \midrule[0.5pt]
  SLSR$^{21}$~\cite{SLSR}                   & ResNet-50~\cite{Resnet}             & $480 \times 480$             & 50.935M                  & 0.787                              & 0.696                                   & 0.080                                & 0.744                             & 0.854              & 0.890              & 0.822                           & 0.030            & 0.841                  & 0.948              & 0.804              & 0.673                           & 0.037            & 0.715                  & 0.892              & 0.840              & 0.765                           & 0.048            & 0.804                  & 0.907              \\
  MGL-R$^{21}$~\cite{COD-MGL}               & ResNet-50~\cite{Resnet}             & $473 \times 473$             & 63.595M                  & 0.775                              & 0.673                                   & 0.088                                & 0.726                             & 0.842              & 0.893              & 0.812                           & 0.030            & 0.834                  & 0.941              & 0.814              & 0.666                           & 0.035            & 0.710                  & 0.890              & 0.833              & 0.739                           & 0.053            & 0.782                  & 0.893              \\
  PFNet$^{21}$~\cite{COD-PFNet}             & ResNet-50~\cite{Resnet}             & $416 \times 416$             & 46.498M                  & 0.782                              & 0.695                                   & 0.085                                & 0.746                             & 0.855              & 0.882              & 0.810                           & 0.033            & 0.828                  & 0.945              & 0.800              & 0.660                           & 0.040            & 0.701                  & 0.890              & 0.829              & 0.745                           & 0.053            & 0.784                  & 0.898              \\
  UJSC$_{\star}^{21}$~\cite{UJSC}           & ResNet-50~\cite{Resnet}             & $480 \times 480$             & 217.982M                 & 0.800                              & 0.728                                   & 0.073                                & 0.772                             & 0.873              & 0.891              & 0.833                           & 0.030            & 0.848                  & 0.955              & 0.809              & 0.684                           & 0.035            & 0.721                  & 0.891              & 0.842              & 0.771                           & 0.046            & 0.806                  & 0.907              \\
  UGTR$^{21}$~\cite{COD-UGTR}               & ResNet-50~\cite{Resnet}             & $473 \times 473$             & 48.868M                  & 0.784                              & 0.684                                   & 0.086                                & 0.735                             & 0.851              & 0.887              & 0.794                           & 0.031            & 0.819                  & 0.940              & 0.817              & 0.666                           & 0.036            & 0.711                  & 0.890              & 0.839              & 0.746                           & 0.052            & 0.787                  & 0.899              \\
  ZoomNet$^{22}$~\cite{COD-ZoomNet}         & ResNet-50~\cite{Resnet}             & $384 \times 384$             & 32.382M                  & 0.820                              & 0.752                                   & 0.066                                & 0.794                             & 0.892              & 0.902              & 0.845                           & \third{0.023}    & 0.864                  & 0.958              & 0.838              & 0.729                           & 0.029            & 0.766                  & 0.911              & 0.853              & 0.784                           & 0.043            & 0.818                  & 0.912              \\
  BSA-Net$^{22}$~\cite{COD-BSA-Net}         & Res2Net-50~\cite{Res2Net}           & $384 \times 384$             & 32.585M                  & 0.794                              & 0.717                                   & 0.079                                & 0.763                             & 0.867              & 0.895              & 0.841                           & 0.027            & 0.858                  & 0.957              & 0.818              & 0.699                           & 0.034            & 0.738                  & 0.901              & 0.842              & 0.771                           & 0.048            & 0.808                  & 0.907              \\
  BGNet$^{22}$~\cite{COD-BGNet}             & Res2Net-50~\cite{Res2Net}           & $416 \times 416$             & 79.853M                  & 0.812                              & 0.749                                   & 0.073                                & 0.789                             & 0.882              & 0.901              & 0.851                           & 0.027            & 0.860                  & 0.954              & 0.831              & 0.722                           & 0.033            & 0.753                  & 0.911              & 0.851              & 0.788                           & 0.044            & 0.820                  & 0.916              \\
  FEDER$^{23}$~\cite{COD-FEDER}             & ResNet-50~\cite{Resnet}             & $384 \times 384$             & 44.129M                  & 0.802                              & 0.738                                   & 0.071                                & 0.781                             & 0.873              & 0.887              & 0.835                           & 0.030            & 0.851                  & 0.954              & 0.822              & 0.716                           & 0.032            & 0.751                  & 0.905              & 0.847              & 0.789                           & 0.044            & 0.824                  & 0.915              \\
  \rowcolor{ours}
  Ours$^{23}$                               & ResNet-50~\cite{Resnet}             & $384 \times 384$             & 28.458M                  & 0.833                              & \third{0.774}                           & 0.065                                & \third{0.813}                     & 0.891              & \third{0.908}      & 0.858                           & \second{0.021}   & 0.874                  & 0.963              & \third{0.861}      & \third{0.768}                   & \third{0.026}    & \third{0.801}          & 0.925              & \third{0.874}      & \third{0.816}                   & \third{0.037}    & \third{0.846}          & \third{0.928}      \\
  \midrule[1pt]
  \multicolumn{24}{c}{\textbf{Vision Transoformer based Methods}}                                                                                                                                                                                                                                                                                                                                                                                                                                                                                                                                                                                                                                  \\
  \midrule[1pt]
  CamoFormer-P$^{22}$~\cite{COD-CamoFormer} & PVTv2-B4~\cite{PVTv2}               & $352 \times 352$             & 71.403M                  & 0.872                              & 0.831                                   & 0.046                                & 0.854                             & 0.938              & 0.910              & 0.865                           & 0.022            & 0.882                  & 0.966              & 0.869              & 0.786                           & 0.023            & 0.811                  & 0.939              & 0.892              & 0.847                           & \third{0.030}    & 0.868                  & \third{0.946}      \\
  \rowcolor{ours}
  Ours$^{23}$                               & PVTv2-B4~\cite{PVTv2}               & $352 \times 352$             & 65.374M                  & \best{0.893}                       & \best{0.862}                            & \best{0.040}                         & \best{0.881}                      & \best{0.949}       & \second{0.929}     & 0.894                           & \third{0.018}    & \second{0.906}         & \second{0.977}     & \second{0.895}     & \third{0.825}                   & \second{0.018}   & \third{0.845}          & \third{0.954}      & \third{0.899}      & 0.859                           & \second{0.029}   & \third{0.879}          & \second{0.949}     \\
  \midrule[0.5pt]
  MSCAF-Net$^{23}$~\cite{COD-MSCAF-Net}     & PVTv2-B2~\cite{PVTv2}               & $352 \times 352$             & 30.364M                  & 0.873                              & 0.828                                   & 0.046                                & 0.852                             & 0.937              & 0.912              & 0.865                           & 0.022            & 0.876                  & 0.970              & 0.865              & 0.775                           & 0.024            & 0.798                  & 0.936              & 0.887              & 0.838                           & 0.032            & 0.860                  & 0.942              \\
  \rowcolor{ours}
  Ours$^{23}$                               & PVTv2-B2~\cite{PVTv2}               & $352 \times 352$             & 28.181M                  & 0.868                              & 0.829                                   & 0.049                                & 0.855                             & 0.926              & 0.916              & 0.876                           & \third{0.018}    & 0.889                  & 0.971              & 0.881              & 0.809                           & 0.020            & 0.834                  & 0.945              & 0.890              & 0.848                           & 0.031            & 0.872                  & 0.941              \\
  \midrule[0.5pt]
  HitNet$^{23}$~\cite{COD-HiTNet}           & PVTv2-B2~\cite{PVTv2}               & $704 \times 704$             & 25.727M                  & 0.849                              & 0.809                                   & 0.055                                & 0.831                             & 0.910              & 0.921              & \third{0.897}                   & 0.019            & 0.900                  & 0.972              & 0.871              & 0.806                           & 0.023            & 0.823                  & 0.938              & 0.875              & 0.834                           & 0.037            & 0.854                  & 0.929              \\
  FSPNet$^{23}$~\cite{COD-FSPNet}           & ViT-B/16~\cite{ViT}                 & $384 \times 384$             & 274.240M                 & 0.856                              & 0.799                                   & 0.050                                & 0.831                             & 0.928              & 0.908              & 0.851                           & 0.023            & 0.867                  & 0.965              & 0.851              & 0.735                           & 0.026            & 0.769                  & 0.930              & 0.878              & 0.816                           & 0.035            & 0.843                  & 0.937              \\
  SARNet$^{23}$~\cite{COD-SARNet}           & PVTv2-B3~\cite{PVTv2}               & $672 \times 672$             & 47.477M                  & 0.874                              & 0.844                                   & 0.046                                & 0.866                             & 0.935              & \best{0.933}       & \best{0.909}                    & \second{0.017}   & \best{0.915}           & \best{0.978}       & 0.885              & 0.820                           & 0.021            & 0.839                  & 0.947              & 0.889              & 0.851                           & 0.032            & 0.872                  & 0.940              \\
  \rowcolor{ours}
  Ours$^{23}$                               & PVTv2-B2~\cite{PVTv2}               & $384 \times 384$             & 28.181M                  & 0.874                              & 0.839                                   & 0.047                                & 0.863                             & 0.931              & 0.922              & 0.884                           & \second{0.017}   & 0.896                  & 0.970              & \third{0.887}      & 0.818                           & \third{0.019}    & 0.841                  & 0.948              & 0.892              & 0.852                           & \third{0.030}    & 0.874                  & 0.943              \\
  \rowcolor{ours}
  Ours$^{23}$                               & PVTv2-B3~\cite{PVTv2}               & $384 \times 384$             & 48.056M                  & 0.885                              & 0.854                                   & \third{0.042}                        & 0.872                             & 0.942              & \third{0.927}      & \second{0.898}                  & \second{0.017}   & \third{0.905}          & \second{0.977}     & \second{0.895}     & \second{0.829}                  & \second{0.018}   & \second{0.848}         & 0.952              & \second{0.900}     & \third{0.861}                   & \best{0.028}     & \second{0.880}         & \second{0.949}     \\
  \rowcolor{ours}
  Ours$^{23}$                               & PVTv2-B4~\cite{PVTv2}               & $384 \times 384$             & 65.374M                  & \third{0.888}                      & \second{0.859}                          & \best{0.040}                         & \second{0.878}                    & \third{0.943}      & 0.925              & \third{0.897}                   & \best{0.016}     & \second{0.906}         & 0.973              & \best{0.898}       & \best{0.838}                    & \best{0.017}     & \best{0.857}           & \second{0.955}     & \second{0.900}     & \best{0.865}                    & \best{0.028}     & \best{0.884}           & \second{0.949}     \\
  \rowcolor{ours}
  Ours$^{23}$                               & PVTv2-B5~\cite{PVTv2}               & $384 \times 384$             & 84.774M                  & \second{0.889}                     & \third{0.857}                           & \second{0.041}                       & \third{0.875}                     & \second{0.945}     & 0.924              & 0.885                           & \third{0.018}    & 0.896                  & \third{0.975}      & \best{0.898}       & 0.827                           & \second{0.018}   & \second{0.848}         & \best{0.956}       & \best{0.903}       & \second{0.863}                  & \best{0.028}     & \best{0.884}           & \best{0.951}       \\
  \bottomrule[2pt]
\end{tabular}

  }
  \label{tab:icod-sota}
\end{table*}

\begin{table*}[t]
  \centering
  \caption{Comparisons of different methods on video COD datasets.
    ``T'' represents the number of frames in the video clip.
    ``T=1'' is the model without temporal difference, where the dynamic routing is not activated.
    ``T=5'' is our final model for the video COD task due to its better performance.
    $\dagger$: For simplicity, the authors' customized CUDA operators are ignored.
    The best three results are highlighted in {\color{reda} \textbf{red}}, {\color{mygreen} \textbf{green}} and {\color{myblue} \textbf{blue}}.
  }
  \resizebox{0.85\linewidth}{!}{%
    \begin{tabular}{l|l|cc|ccccccc|ccccccc}
  \toprule[2pt]
  \rowcolor{tabtitle}
                                              &                            &                       &                          & \multicolumn{7}{c|}{\textbf{CAD}} & \multicolumn{7}{c}{\textbf{MoCA-Mask-TE}}                                                                                                                                                                                                                                                        \\
  \rowcolor{tabtitle}
  \multirow{-2}{*}{Model}                     & \multirow{-2}{*}{Backbone} & \multirow{-2}{*}{FPS} & \multirow{-2}{*}{Param.} & S$_{m}~\uparrow$                  & F$^{\omega}_{\beta}~\uparrow$             & MAE$~\downarrow$ & F$_{\beta}~\uparrow$ & E$_{m}~\uparrow$ & mDice$~\uparrow$ & mIoU$~\uparrow$ & S$_{m}~\uparrow$ & F$^{\omega}_{\beta}~\uparrow$ & MAE$~\downarrow$ & F$_{\beta}~\uparrow$ & E$_{m}~\uparrow$ & mDice$~\uparrow$ & mIoU$~\uparrow$ \\
  \midrule[1pt]
  EGNet$^{19}$~\cite{EGNet}                   & ResNet-50~\cite{Resnet}    & 5.682                 & 111.693M                 & 0.619                             & 0.298                                     & 0.044            & 0.350                & 0.666            & 0.324            & 0.243           & 0.547            & 0.110                         & 0.035            & 0.136                & 0.574            & 0.143            & 0.096           \\
  BASNet$^{19}$~\cite{BASNet}                 & ResNet-50~\cite{Resnet}    & 37.361                & 87.060M                  & 0.639                             & 0.349                                     & 0.054            & 0.394                & 0.773            & 0.393            & 0.293           & 0.561            & 0.154                         & 0.042            & 0.173                & 0.598            & 0.190            & 0.137           \\
  CPD$^{19}$~\cite{CPD}                       & ResNet-50~\cite{Resnet}    & 43.710                & 47.851M                  & 0.622                             & 0.289                                     & 0.049            & 0.357                & 0.667            & 0.330            & 0.239           & 0.561            & 0.121                         & 0.041            & 0.152                & 0.613            & 0.162            & 0.113           \\
  PraNet$^{20}$~\cite{PraNet}                 & ResNet-50~\cite{Resnet}    & 41.886                & 32.547M                  & 0.629                             & 0.352                                     & 0.042            & 0.397                & 0.763            & 0.378            & 0.290           & 0.614            & 0.266                         & 0.030            & 0.296                & 0.674            & 0.311            & 0.234           \\
  SINet$^{20}$~\cite{COD10K}                  & ResNet-50~\cite{Resnet}    & 56.509                & 48.947M                  & 0.636                             & 0.346                                     & 0.041            & 0.395                & 0.775            & 0.381            & 0.283           & 0.598            & 0.231                         & 0.028            & 0.256                & 0.699            & 0.277            & 0.202           \\
  SINet-V2$^{21}$~\cite{SINetV2}              & Res2Net-50~\cite{Res2Net}  & 38.307                & 26.976M                  & 0.653                             & 0.382                                     & 0.039            & 0.432                & 0.762            & 0.413            & 0.318           & 0.588            & 0.204                         & 0.031            & 0.229                & 0.642            & 0.245            & 0.180           \\
  PNS-Net$^{21}$~\cite{VideoPolypSeg-PNS-Net} & ResNet-50~\cite{Resnet}    & 142.889               & 26.874M$^\dagger$        & 0.655                             & 0.325                                     & 0.048            & 0.417                & 0.673            & 0.384            & 0.290           & 0.526            & 0.059                         & 0.035            & 0.084                & 0.530            & 0.084            & 0.054           \\
  RCRNet$^{19}$~\cite{VCODRelated-RCRNet}     & ResNet-50~\cite{Resnet}    & 58.132                & 53.790M                  & 0.627                             & 0.287                                     & 0.048            & 0.328                & 0.666            & 0.309            & 0.229           & 0.555            & 0.138                         & 0.033            & 0.159                & 0.527            & 0.171            & 0.116           \\
  MG$^{21}$~\cite{VCODRelated-MG}             & VGG-style CNN              & 272.948               & 4.766M                   & 0.594                             & 0.336                                     & 0.059            & 0.375                & 0.692            & 0.368            & 0.268           & 0.530            & 0.168                         & 0.067            & 0.195                & 0.561            & 0.181            & 0.127           \\
  STL-Net-LT$^{22}$~\cite{VCOD-MoCA-Mask}     & PVTv2-B5~\cite{PVTv2}      & 35.646                & 82.303M$^\dagger$        & 0.696                             & 0.481                                     & 0.030            & 0.524                & \third{0.845}    & 0.493            & 0.402           & 0.631            & 0.311                         & 0.027            & 0.331                & \best{0.759}     & 0.360            & 0.272           \\
  STL-Net-ST$^{22}$~\cite{VCOD-MoCA-Mask}     & PVTv2-B5~\cite{PVTv2}      & 6.488                 & 82.383M                  & 0.696                             & 0.471                                     & 0.031            & 0.515                & 0.827            & 0.484            & 0.392           & 0.637            & 0.304                         & 0.027            & 0.328                & \third{0.734}    & 0.356            & 0.271           \\
  \midrule[0.5pt]
  Ours$^{23}$ (T=1)                           & PVTv2-B5~\cite{PVTv2}      & 11.741                & 84.774M                  & 0.721                             & 0.525                                     & \third{0.024}    & 0.567                & 0.759            & 0.523            & 0.436           & 0.690            & 0.395                         & \third{0.017}    & \third{0.424}        & 0.702            & \third{0.420}    & 0.353           \\
  Ours$^{23}$ (T=3)                           & PVTv2-B5~\cite{PVTv2}      & 23.207                & 84.775M                  & \third{0.741}                     & \third{0.565}                             & \second{0.021}   & \third{0.598}        & \second{0.849}   & \third{0.571}    & \third{0.487}   & \third{0.703}    & \third{0.425}                 & \best{0.010}     & \second{0.444}       & 0.717            & \second{0.446}   & \third{0.378}   \\
  \rowcolor{ours}
  Ours$^{23}$ (T=5)                           & PVTv2-B5~\cite{PVTv2}      & 26.454                & 84.776M                  & \best{0.757}                      & \best{0.593}                              & \best{0.020}     & \best{0.631}         & \best{0.865}     & \best{0.599}     & \best{0.510}    & \best{0.734}     & \best{0.476}                  & \best{0.010}     & \best{0.497}         & \second{0.736}   & \best{0.497}     & \best{0.422}    \\
  Ours$^{23}$ (T=7)                           & PVTv2-B5~\cite{PVTv2}      & 30.292                & 84.778M                  & \second{0.751}                    & \second{0.583}                            & \best{0.020}     & \second{0.618}       & 0.837            & \second{0.591}   & \second{0.505}  & \second{0.706}   & \second{0.428}                & \second{0.011}   & \second{0.444}       & 0.717            & \second{0.446}   & \second{0.382}  \\
  \bottomrule[2pt]
\end{tabular}

  }
  \label{tab:vcod-sota}
\end{table*}

\subsection{Loss Functions}\label{sec:loss}

\parhead{Binary Cross Entropy Loss (BCE)}
The BCE loss function is widely used in various binary image segmentation tasks and its mathematical form is:
$l_{BCE}^{i,j} = - \mathbf{g}_{i,j} \log \mathbf{p}_{i,j} - (1 - \mathbf{g}_{i,j}) \log (1 - \mathbf{p}_{i,j})$, where $\mathbf{g}_{i,j} \in \{0, 1\}$ and $\mathbf{p}_{i,j} \in [0, 1]$ denote the ground truth and the predicted value at position $(i,j)$, respectively.
As shown in Fig.~\ref{fig:visualation}, due to the complexity of the COD data, if trained only under the BCE, the model produces serious ambiguity and uncertainty in the prediction and fails to accurately capture objects, of which both will reduce the reliability of COD.

\parhead{\myUaL~(\myUAL)}
To force the model to enhance ``confidence'' in decision-making and increase the penalty for fuzzy prediction, we design a strong constraint as the auxiliary of the BCE, \ie the \myual (\myUAL).
In the final probability map of the camouflaged object, the pixel value range is $[0, 1]$, where $0$ means the pixel belongs to the background, and $1$ means it belongs to the camouflaged object.
Therefore, the closer the predicted value is to $0.5$, the more uncertain the determination about the property of the pixel is.
To optimize it, a direct way is to use ambiguity as the supplementary loss for these difficult samples.
To this end, we first need to define the ambiguity measure of the pixel $x$, which maximizes at $x=0.5$ and minimizes at $x=0$ or $x=1$.
As a loss, the function should be smooth and continuous with only a finite number of non-differentiable points.
For brevity, we empirically consider two forms, $\Phi_{pow}^{\alpha}(x)=1-|2x-1|^{\alpha}$ based on the power function and $\Phi_{exp}^{\alpha}(x)=e^{-(\alpha (x-0.5))^{2}}$ based on the exponential function.
Besides, inspired by the form of the weighted BCE loss, we also try to use $\alpha = 1+\Phi_{pow}^{2}(x)$ as the weight of BCE loss to increase the loss of hard pixels.
The different forms and corresponding results are listed in Fig.~\ref{fig:loss_form} and Tab.~\ref{tab:loss_form}.
After massive experiments (Sec.~\ref{sec:loss_form}), the proposed \myUAL is formulated as
\begin{equation}
  \begin{split}
    l_{\myUAL}^{i,j} = 1 - \Delta_{i,j} = 1 - |2\mathbf{p}_{i,j}-1|^{2},
  \end{split}
  \label{equ:ual}
\end{equation}
where $\Delta$ represents the certainty of the prediction.

\begin{figure*}[!t]
  \centering
  \subfloat%
  {\centering
    \includegraphics[width=\linewidth]{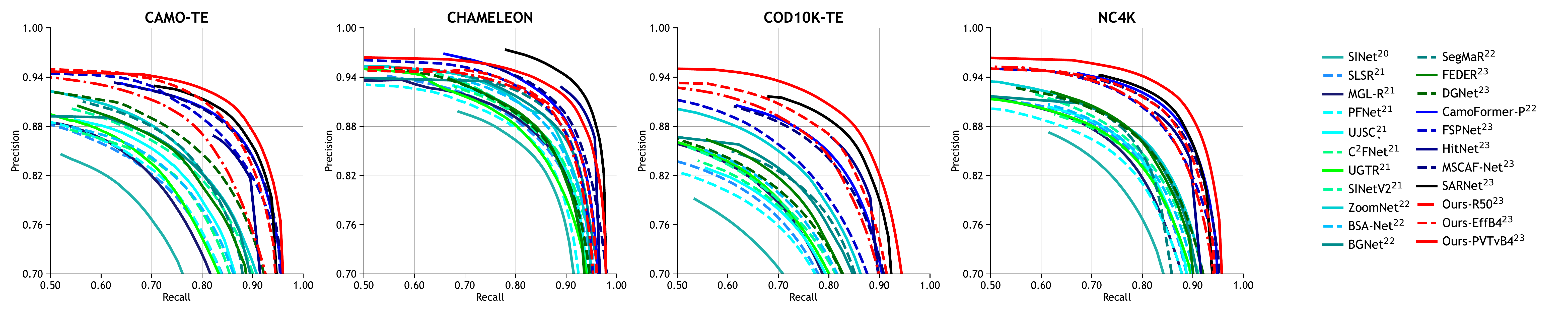}
  }
  \\
  \vspace{-1.5em}
  \subfloat%
  {\centering
    \includegraphics[width=\linewidth]{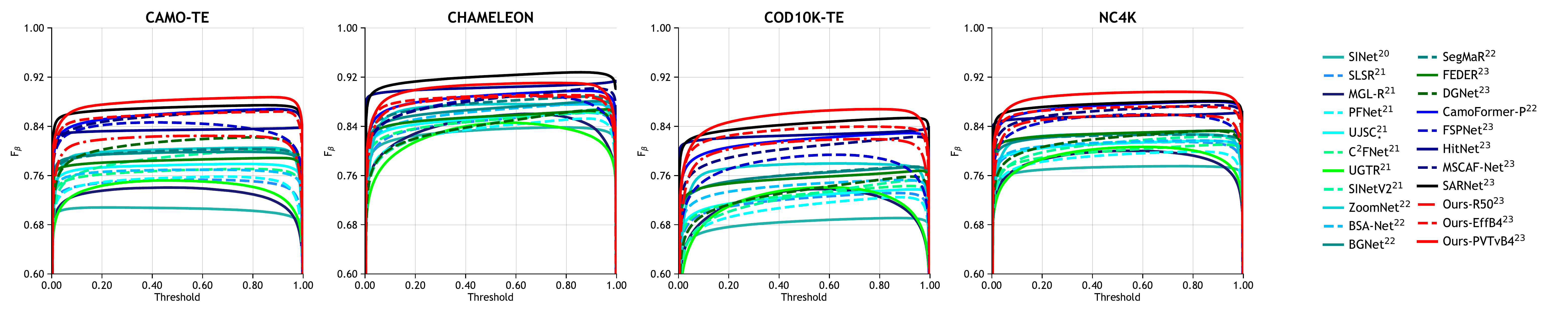}
  }
  \\
  \vspace{-1.5em}
  \subfloat%
  {\centering
    \includegraphics[width=\linewidth]{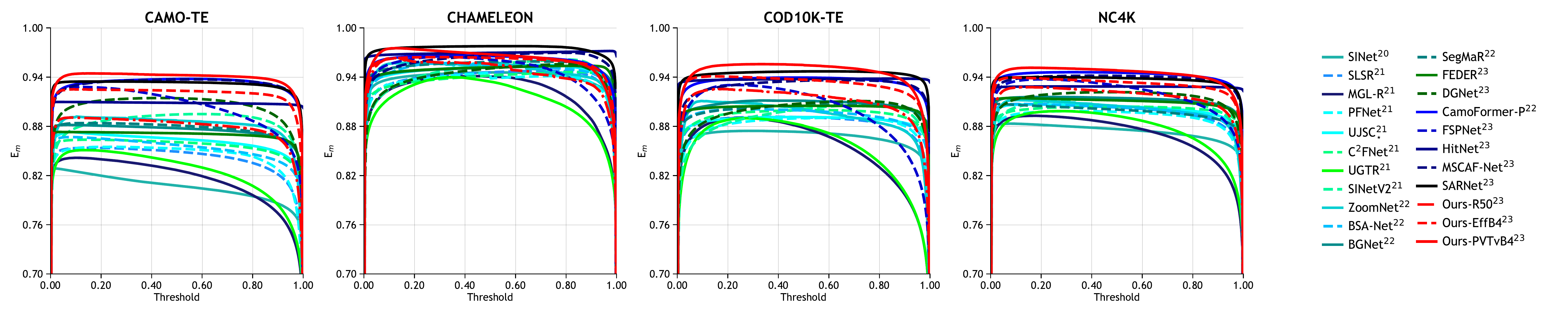}
  }
  \caption{PR, $F_{\beta}$ and $E_{m}$ curves of the proposed model and recent SOTA algorithms over four COD datasets.}
  \label{fig:prfmem}
\end{figure*}

\parhead{Total Loss Function}
Finally, the total loss function can be written as $L = L_{BCE} + \lambda L_{\myUAL}$, where $\lambda$ is the balance coefficient. We design three adjustment strategies of $\lambda$, \ie a fixed constant value, an increasing linear strategy, and an increasing cosine strategy in Sec.~\ref{sec:loss_form}.
From the results shown in Tab.~\ref{tab:exp_losscoef}, we find that the increasing strategies, especially ``cosine'', do achieve better performance.
So, the cosine strategy is used by default.

\begin{figure*}[!h]
  \centering
  \subfloat[Some recent image COD methods and ours on different types of samples.]
  {\centering
    \includegraphics[width=\linewidth]{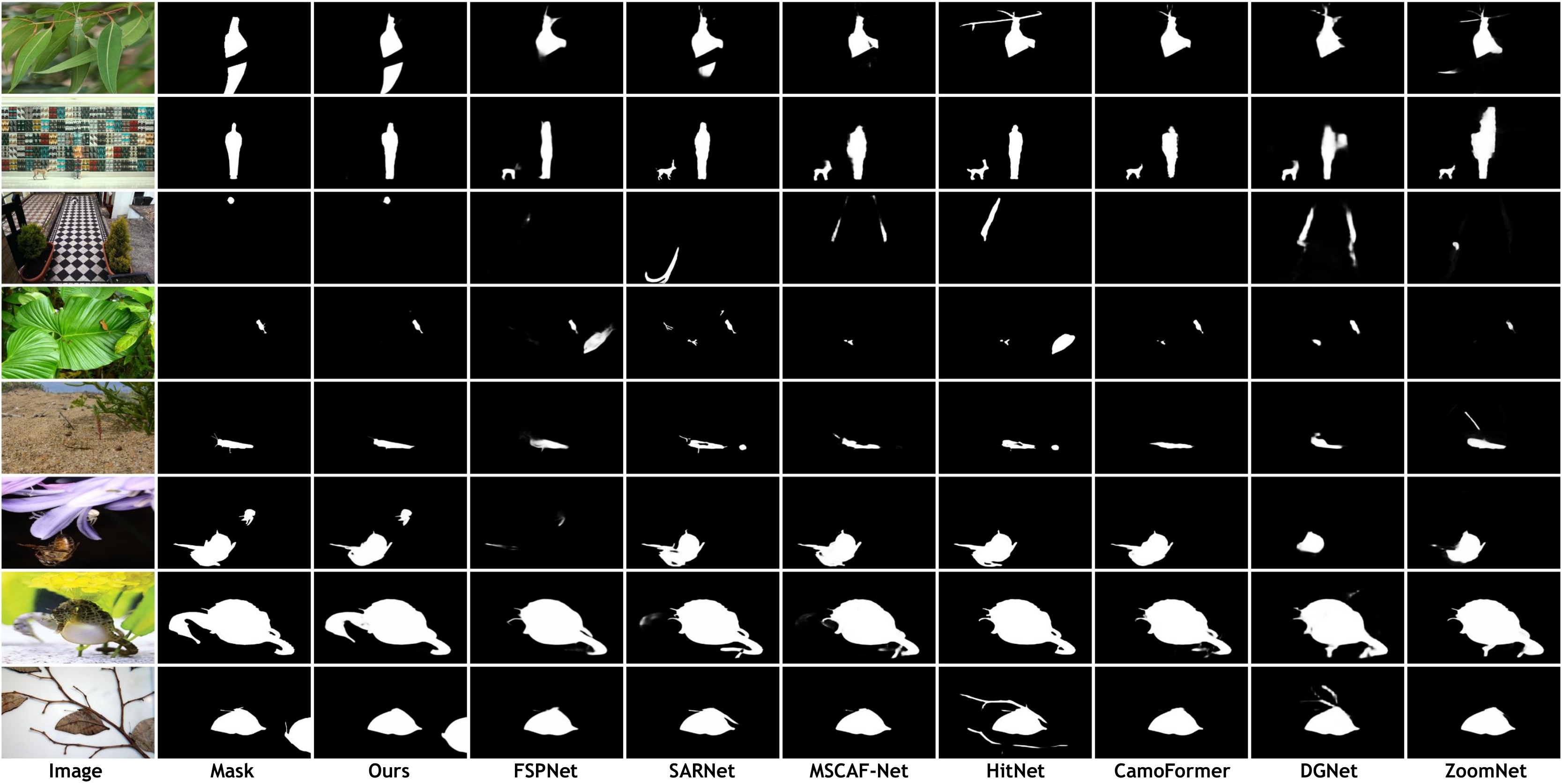}
    \label{fig:icod-viscmp}
  }
  \\
  \vspace{-1em}
  \subfloat[``Ours (T=5)'' and recent best STL-Net-LT~\cite{VCOD-MoCA-Mask} for video COD. For better visualization, grey-scale predictions are thresholded by 0.5 here.]
  {\centering
    \includegraphics[width=\linewidth]{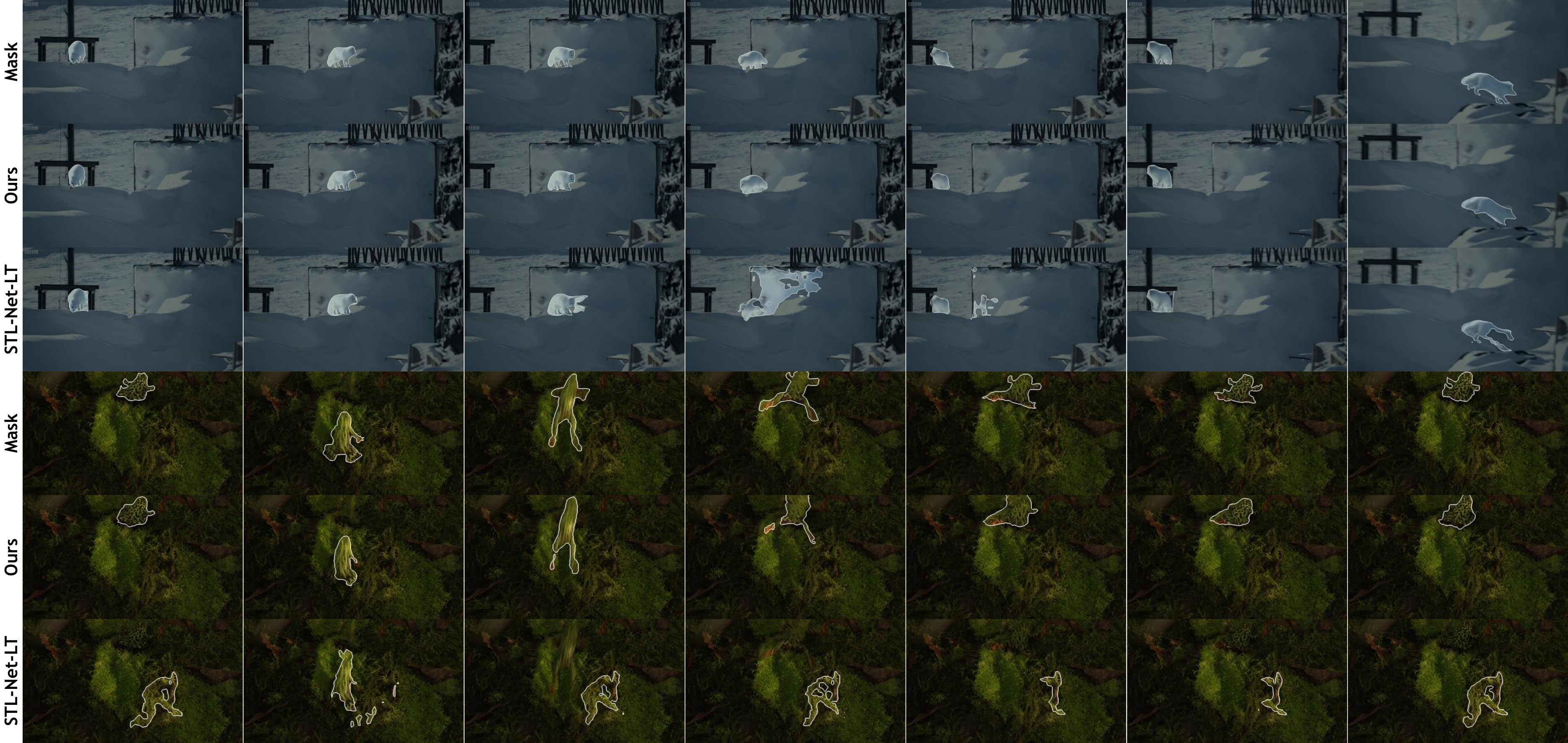}
    \label{fig:vcod-viscmp}
  }
  \caption{Visual comparisons of image and video COD methods.}
  \label{fig:viscmp}
\end{figure*}

\section{Experiments}

\subsection{Experiment Setup}

\subsubsection{Datasets}

We use four image COD datasets,
CAMO~\cite{CAMO},
CHAMELEON~\cite{CHAMELEON},
COD10K~\cite{COD10K}
and NC4K~\cite{SLSR},
and two video COD datasets,
MoCA-Mask~\cite{VCOD-MoCA-Mask}
and CAD~\cite{VCOD-CAD}.

\parhead{Image COD}
CAMO consists of 1,250 camouflaged and 1,250 non-camouflaged images.
CHAMELEON contains 76 hand-annotated images.
COD10K includes 5,066 camouflaged, 3,000 background, and 1,934 non-camouflaged images.
NC4K is another large-scale COD testing dataset including 4,121 images from the Internet.
Following the data partition of~\cite{COD10K, COD-PFNet, COD-ZoomNet, COD-CamoFormer, COD-DGNet}, we use all images with camouflaged objects in the experiments, in which 3,040 images from COD10K and 1,000 images from CAMO for training, and the rest ones for testing.

\parhead{Video COD}
The recent MoCA-Mask is reorganized from the moving camouflaged animals (MoCA) dataset~\cite{VCOD-MoCA}.
MoCA-Mask includes 71 sequences with 19,313 frames for training, and 16 sequences with 3,626 frames for testing, respectively.
And it extends the original bounding box annotations of the MoCA dataset to finer segmentation annotations.
The camouflaged animal dataset, \ie CAD, includes 9 short video sequences in total that accompany 181 hand-labeled masks on every $5^{th}$ frame.
Except for the training set containing 71 frame sequences of MoCA-Mask for training, the rest of the video sequences are used for testing as~\cite{VCOD-MoCA-Mask}.

\subsubsection{Evaluation Criteria}

For image COD, we use eight common metrics for evaluation based on~\cite{PySODMetrics, PySODEvalToolkit}, including
S-measure~\cite{Smeasure} (S$_{m}$),
weighted F-measure~\cite{wFmeasure} (F$_{\beta}^{\omega}$),
mean absolute error (MAE),
F-measure~\cite{Fmeasure} (F$_{\beta}$),
E-measure~\cite{Emeasure} (E$_{m}$),
precision-recall curve (PR curve),
F$_{\beta}$-threshold curve (F$_{\beta}$ curve),
and E$_{m}$-threshold curve (E$_{m}$ curve).
And for video COD, the seven metrics used in the recent pioneer work~\cite{VCOD-MoCA-Mask} are introduced to evaluate existing methods, including S$_{m}$, F$_{\beta}^{\omega}$, MAE, F$_{\beta}$, E$_{m}$, mDice, and mIoU.

\subsubsection{Implementation Details}

The proposed camouflaged object detector is implemented with PyTorch~\cite{PyTorch}.
And the training settings are on par with recent best practices~\cite{COD10K, COD-PFNet, COD-ZoomNet, COD-DGNet, COD-CamoFormer, VCOD-MoCA-Mask}.
The encoder is initialized with the parameters of ResNet~\cite{Resnet}, EfficientNet~\cite{EfficientNet} and PVTv2~\cite{PVTv2} pretrained on ImageNet, and the remaining parts are randomly initialized.
The optimizer Adam with \texttt{betas} = (0.9, 0.999) is chosen to update the model parameters.
The learning rate is initialized to 0.0001 and follows a \texttt{step} decay strategy.
The entire model is trained for 150 epochs with a batch size of 8 in an end-to-end manner on an NVIDIA 2080Ti GPU.
During training, the main input $\mathbf{I}^{1.0}$ and the ground truth $\mathbf{G}$ are bi-linearly interpolated to $384 \times 384$.
At inference, $\mathbf{I}^{1.0}$ and the prediction are interpolated to $384 \times 384$ and the size of $\mathbf{G}$, respectively.
Considering that some methods use a smaller resolution $352 \times 352$, we list the results with the corresponding resolution and backbone in Tab.~\ref{tab:icod-sota}.
Random flipping, rotating, and color jittering are employed to augment the training data and no test-time augmentation is used during inference.
In addition, for a fair comparison, we follow the settings used in~\cite{VCOD-MoCA-Mask} in the video COD task, \ie pretraining on COD10K-TR + fine-tuning on MoCA-Mask-TR.

\subsection{Comparisons with State-of-the-arts}

In order to evaluate the proposed method, we construct a comparison of it with several recent state-of-the-art image-based and video-based methods.
The results of all these methods come from existing public data or are generated by models retrained with the corresponding code.
The comparison of the proposed method is summarized in Tab.~\ref{tab:icod-sota}, Tab.~\ref{tab:vcod-sota}, and Fig.~\ref{fig:prfmem}, and the qualitative performance of all methods is shown in Fig.~\ref{fig:icod-viscmp} and Fig.~\ref{fig:vcod-viscmp}.

\parhead{Quantitative Evaluation}
Tab.~\ref{tab:icod-sota} and Tab.~\ref{tab:vcod-sota} show the detailed comparison results for image and video COD tasks, respectively.
It can be seen that the proposed model consistently and significantly surpasses recent methods on all datasets without relying on any post-processing tricks.
Compared with the recent several best methods with different backbones, ZoomNet~\cite{COD-ZoomNet}, DGNet~\cite{COD-DGNet} and SARNet~\cite{COD-SARNet} for image COD and STL-Net-LT~\cite{VCOD-MoCA-Mask} for video COD, although they have suppressed other existing methods, our method still shows the obvious improvement over the conference version~\cite{COD-ZoomNet} and the performance advantages over recent methods on these datasets.
Besides, PR, F$_{\beta}$ and E$_{m}$ curves marked by the red color in Fig.~\ref{fig:prfmem} also demonstrate the effectiveness of the proposed method.
The flatness of the F$_{\beta}$ and E$_{m}$ curves reflects the consistency and uniformity of the prediction.
Our curves are almost horizontal, which can be attributed to the effect of the proposed \myUAL.
It drives the predictions to be more polarized and reduces the ambiguity.

\parhead{Qualitative Evaluation}
Visual comparisons of different methods on several typical samples are shown in Fig.~\ref{fig:icod-viscmp} and Fig.~\ref{fig:vcod-viscmp}.
They present the complexity in different aspects, such as
small objects (Row 3-5),
middle objects (Row 1 and 2),
big objects (Row 6-8),
occlusions (Row 1 and 6),
background interference (Row 1-3 and 6-8),
and indefinable boundaries (Row 2, 3, and 5) in Fig.~\ref{fig:icod-viscmp},
and motion blur in Fig.~\ref{fig:vcod-viscmp}.
These results intuitively show the superior performance of the proposed method.
In addition, it can be noticed that our predictions have clearer and more complete object regions, sharper contours, and better temporal consistency.

\subsection{Ablation Studies}\label{ablation}

In this section, we perform comprehensive ablation analyses on different components.
Because COD10K is the most widely-used large-scale COD dataset, and contains various objects and scenes, all subsequent ablation experiments are carried out on it.

\subsubsection{Effectiveness of Proposed Modules}
\label{sec:ablation_siuhmu}

\begin{figure}[t]
  \centering
  \includegraphics[width=\linewidth]{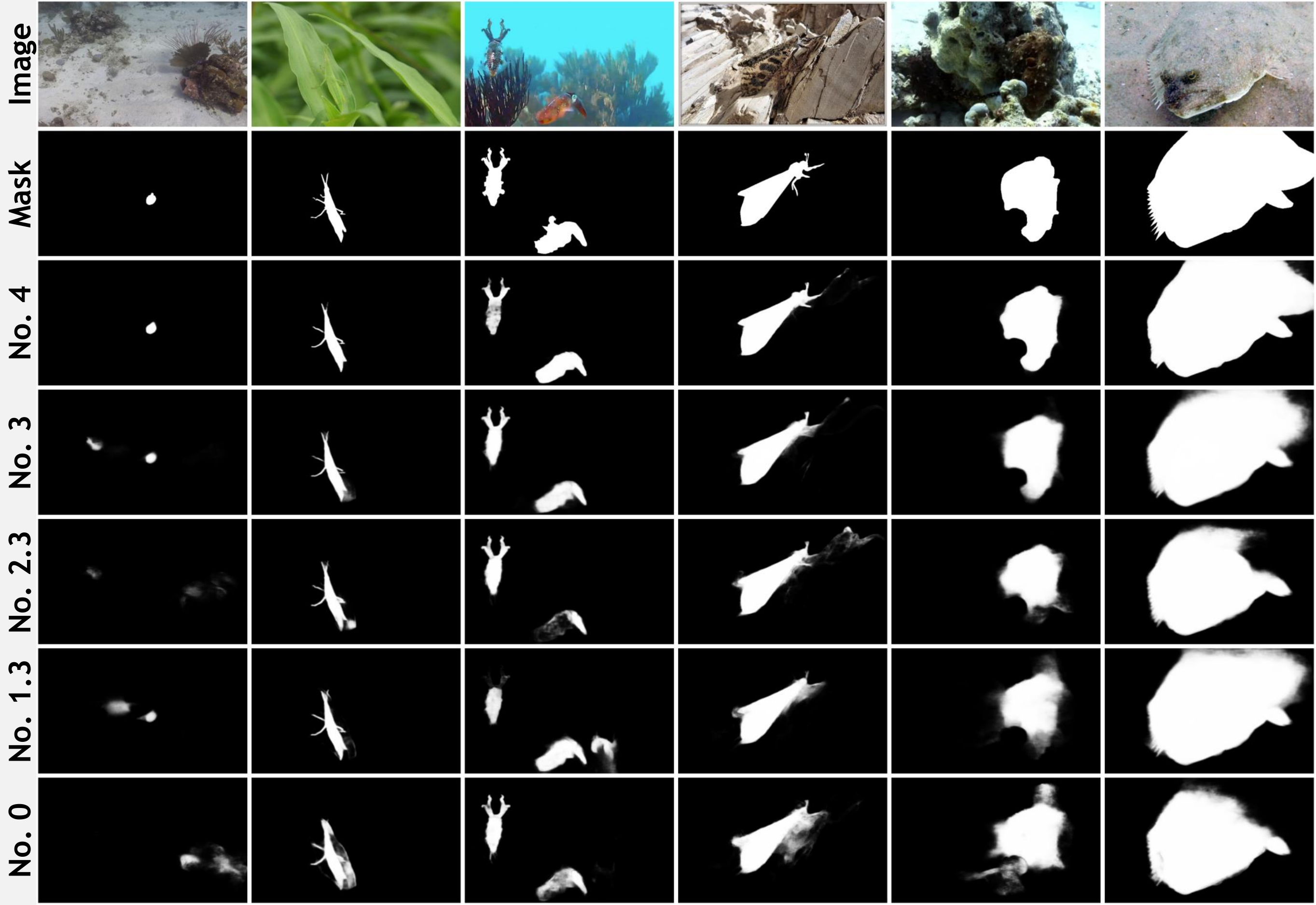}
  \caption{Visual comparisons for showing the effects of the proposed components. The names ``No.$\star$'' are the same as those in Tab.~\ref{tab:ablationstudy}. Please zoom in to see more details.}
  \label{fig:visualation}
\end{figure}

In the proposed model, both the \myMHSIU and the \myRGPU are very important structures.
We install them one by one on the baseline model to evaluate their performance.
The results are shown in Tab.~\ref{tab:ablationstudy}.
Note that only the inputs of the main scale are used in our baseline ``0'' and the models ``1.x''.
As can be seen, our baseline shows a good performance, probably due to the more reasonable network architecture.
From the table, it can be seen that the two proposed modules make a significant contribution to the performance when compared to the baseline.
Besides, the visual results in Fig.~\ref{fig:visualation} show that the two modules can benefit each other and reduce their errors to locate and distinguish objects more accurately.
These components effectively help the model to excavate and distill the critical and valuable semantics and improve the capability of distinguishing hard objects.
Under the cooperation between the proposed modules and the proposed loss function, \myModel can completely capture the camouflaged objects of different scales and generate the predictions with higher consistency.

\begin{table*}[!t]
  \centering
  \caption{Ablation study on image COD datasets.
    \myRGPU: \myRgpu;
    \myMHSIU: \myMhsiu;
    $L_{UAL}$: \myUal;
    ARG: Average relative gain;
    \none: Unable to converge.}
  \resizebox{0.7\linewidth}{!}{%
    \begin{tabular}{l|l|cc|ccccc|ccccc|c}
  \toprule[2pt]
  \rowcolor{tabtitle}
                        &                                              &                              &                          & \multicolumn{5}{c|}{\textbf{COD10K}} & \multicolumn{5}{c|}{\textbf{NC4K}} &                                                                                                                                                                                                                     \\
  \rowcolor{tabtitle}
  \multirow{-2}{*}{No.} & \multirow{-2}{*}{Model}                      & \multirow{-2}{*}{Param. (M)} & \multirow{-2}{*}{GFLOPs} & S$_{m}$ $\uparrow$                   & F$^{\omega}_{\beta}$ $\uparrow$    & MAE $\downarrow$ & F$_{\beta}$ $\uparrow$ & E$_{m}$ $\uparrow$ & S$_{m}$ $\uparrow$ & F$^{\omega}_{\beta}$ $\uparrow$ & MAE $\downarrow$ & F$_{\beta}$ $\uparrow$ & E$_{m}$ $\uparrow$ & \multirow{-2}{*}{$\Delta$} \\
  \midrule[1pt]
  0                     & Baseline                                     & 25.216                       & 41.000                   & 0.826                                & 0.677                              & 0.035            & 0.724                  & 0.903              & 0.850              & 0.756                           & 0.049            & 0.800                  & 0.912              & 0.00\%                     \\
  \midrule[0.5pt]
  1.1                   & No.0 + RGPU ($G=2$)                          & 25.583                       & 47.801                   & 0.836                                & 0.691                              & 0.033            & 0.739                  & 0.913              & 0.859              & 0.769                           & 0.047            & 0.808                  & 0.916              & $\uparrow$2.05\%           \\
  1.2                   & No.0 + RGPU ($G=4$)                          & 26.364                       & 62.748                   & 0.839                                & 0.696                              & 0.032            & 0.740                  & 0.915              & 0.859              & 0.770                           & 0.046            & 0.809                  & 0.916              & $\uparrow$2.71\%           \\
  1.3                   & No.0 + RGPU ($G=6$)                          & 27.145                       & 77.694                   & 0.843                                & 0.705                              & 0.031            & 0.749                  & 0.914              & 0.863              & 0.778                           & 0.045            & 0.816                  & 0.919              & $\uparrow$3.77\%           \\
  1.4                   & No.0 + RGPU ($G=8$)                          & 27.926                       & 92.641                   & 0.839                                & 0.700                              & 0.032            & 0.744                  & 0.911              & 0.861              & 0.775                           & 0.045            & 0.812                  & 0.915              & $\uparrow$3.10\%           \\
  1.5                   & No.0 + Res2Net Block                         & 27.248                       & 80.936                   & 0.833                                & 0.685                              & 0.033            & 0.732                  & 0.910              & 0.854              & 0.761                           & 0.048            & 0.801                  & 0.913              & $\uparrow$1.30\%           \\
  \midrule[0.5pt]
  2.1                   & No.0 + SIU                                   & 26.762                       & 152.720                  & 0.852                                & 0.721                              & 0.031            & 0.763                  & 0.919              & 0.862              & 0.780                           & 0.044            & 0.818                  & 0.919              & $\uparrow$4.60\%           \\
  2.2                   & No.0 + MH-SIU ($M=2$)                        & 26.576                       & 150.306                  & 0.853                                & 0.724                              & 0.030            & 0.768                  & 0.922              & 0.864              & 0.784                           & 0.043            & 0.822                  & 0.922              & $\uparrow$5.41\%           \\
  2.3                   & No.0 + MH-SIU ($M=4$)                        & 26.530                       & 149.100                  & 0.856                                & 0.728                              & 0.029            & 0.770                  & 0.923              & 0.870              & 0.789                           & 0.043            & 0.825                  & 0.924              & $\uparrow$6.03\%           \\
  2.4                   & No.0 + MH-SIU ($M=8$)                        & 26.518                       & 148.498                  & 0.856                                & 0.726                              & 0.029            & 0.769                  & 0.924              & 0.867              & 0.783                           & 0.044            & 0.820                  & 0.922              & $\uparrow$5.59\%           \\
  \midrule[0.5pt]
  3                     & No.0 + RGPU ($G=6$) +MHSIU ($M=4$)           & 28.458                       & 185.794                  & 0.861                                & 0.734                              & 0.029            & 0.773                  & 0.924              & 0.875              & 0.792                           & 0.042            & 0.826                  & 0.925              & $\uparrow$6.56\%           \\
  \rowcolor{ours}
  4                     & No.3 + $L_{UAL}$                             & 28.458                       & 185.794                  & 0.861                                & 0.768                              & 0.026            & 0.801                  & 0.925              & 0.874              & 0.816                           & 0.037            & 0.846                  & 0.928              & $\uparrow$9.94\%           \\
  4.1                   & No.3 + $L_{ADB}$~\cite{A2Sv1}                & 28.458                       & 185.794                  & 0.855                                & 0.752                              & 0.027            & 0.787                  & 0.923              & 0.869              & 0.803                           & 0.040            & 0.836                  & 0.926              & $\uparrow$8.12\%                           \\
  4.2                   & No.3 + $L_{CSD}$~\cite{A2Sv2}                & 28.458                       & 185.794                  & 0.854                                & 0.757                              & 0.028            & 0.789                  & 0.923              & 0.865              & 0.802                           & 0.040            & 0.835                  & 0.925              & $\uparrow$7.84\%                           \\
  4.3                   & No.3 + $L_{BTM}$~\cite{A2Sv2}                & 28.458                       & 185.794                  & \none                                & \none                              & \none            & \none                  & \none              & \none              & \none                           & \none            & \none                  & \none              & \none                      \\
  4.4                   & No.3 + $L_{MS}$~\cite{A2Sv2}                 & 28.458                       & 185.794                  & 0.861                                & 0.739                              & 0.028            & 0.778                  & 0.923              & 0.870              & 0.790                           & 0.042            & 0.825                  & 0.923              & $\uparrow$6.85\%                           \\
  4.5                   & No.3 + $L_{CSD}+L_{BTM}+L_{MS}$~\cite{A2Sv2} & 28.458                       & 185.794                  & \none                                & \none                              & \none            & \none                  & \none              & \none              & \none                           & \none            & \none                  & \none              & \none                      \\
  \bottomrule[2pt]
\end{tabular}%

  }
  \label{tab:ablationstudy}
\end{table*}

\begin{figure}[t]
  \centering
  \includegraphics[width=\linewidth]{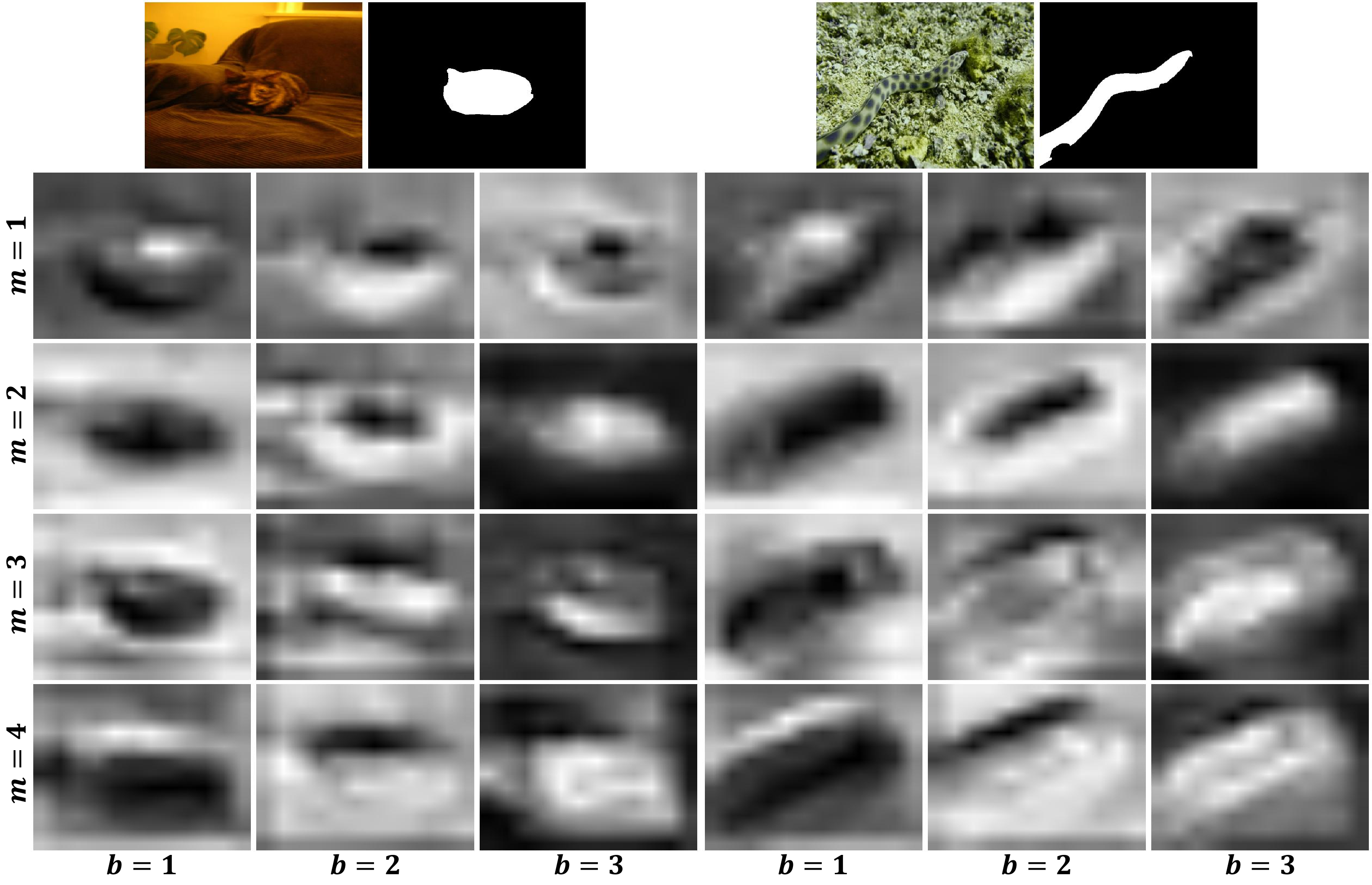}
  \caption{Spatial attention patterns from different heads $\{m=i\}^M_{i=1}$ corresponding to all branches $\{b=i\}^3_{i=1}$ in the deepest \myMHSIU.}
  \label{fig:siuattn}
\end{figure}

\begin{figure}[t]
  \centering
  \includegraphics[width=\linewidth]{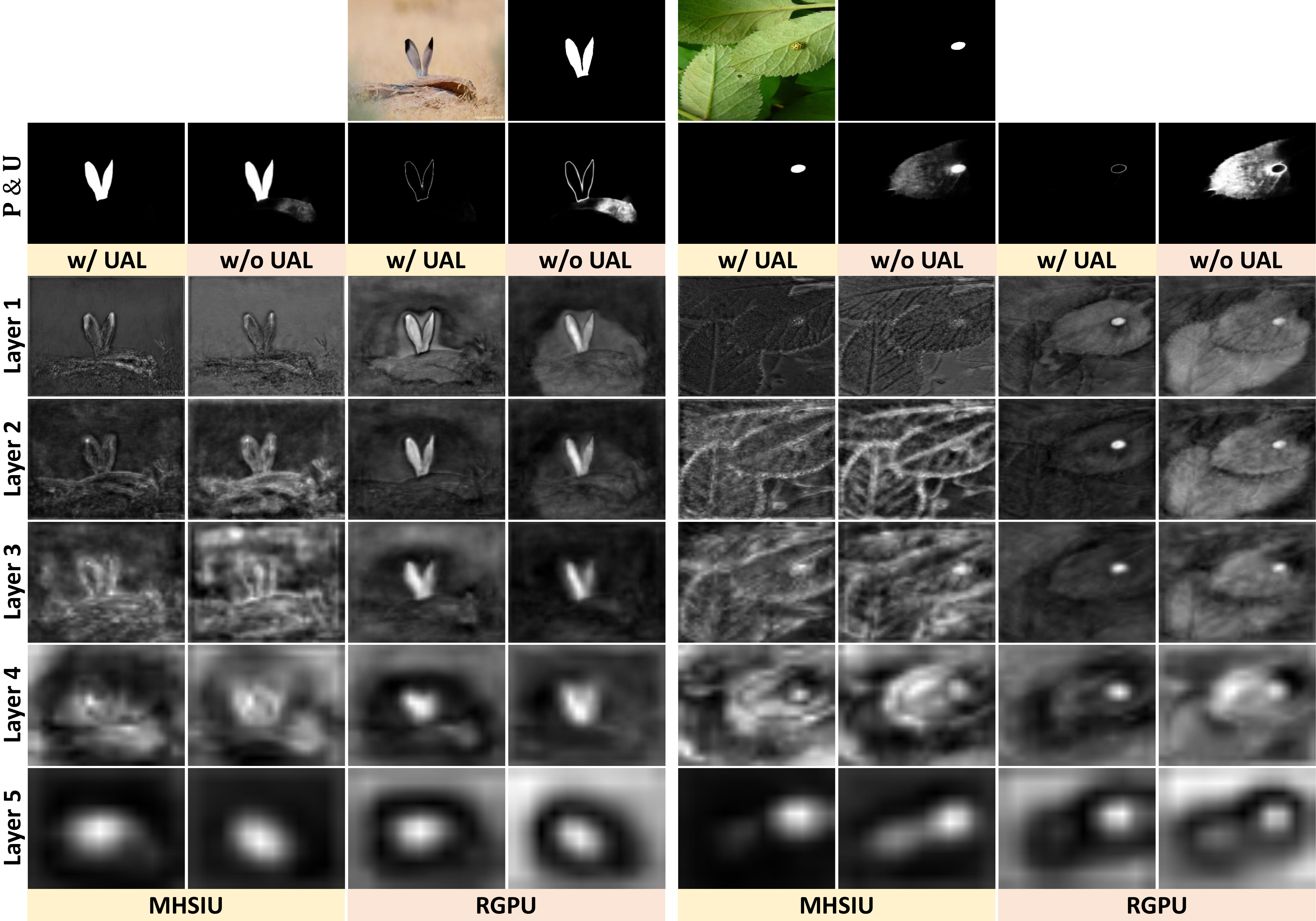}
  \caption{Intermediate feature maps from different stages of the scale merging subnetwork and the hierarchical difference propagation decoder for showing the effects of the proposed components. $\mathbf{P}$: Prediction of the model; $\mathbf{U}$: Uncertainty map generated by $\Phi_{pow}^{2}(x)$ as stated in Sec.~\ref{sec:loss_form}.}
  \label{fig:featvisualablation}
\end{figure}

\subsubsection{Groups in \myRGPU and Heads in \myMHSIU}
\label{sec:num_groups}

\parhead{\myRGPU}
The recursive iteration strategy in the proposed \myRGPU stimulates the potential semantic diversity of feature representation.
The results in Tab.~\ref{tab:ablationstudy} show the impact of different number of iteration groups on performance.
\textit{Although increasing the number of groups introduces more parameters and computation, the performance does not keep increasing.}
It can be seen from the results that the best performance appears when the number of groups is equal to 6 and also achieves a good balance between performance and efficiency.
So we use 6 groups as the default setting in all \myRGPUs.
Besides, we also replace the \myRGPU with the Res2Net~\cite{Res2Net} block while ensuring consistent parameters and computations.
As seen from the results in Tab.~\ref{tab:ablationstudy}, our \myRGPU exhibits better performance.
This can be attributed to the group-adaptive channel-wise modulation, which provides more efficient information modeling.

\begin{table*}[t]
  \centering
  \caption{
    Ablation study on video COD datasets. ``w/ difference image'' and ``w/ optical flow'': Using the difference image and optical flow to introduce the motion cues and replace the our temporal components.
    ``w/o intra-frame self-attention'', ``w/o cross-frame cues diffusion'', and  ``w/o temporal shifting'': Removing the corresponding operation in the \myRGPU, where ``w/o temporal shifting'' invalidates the conditional temporal modeling capability of our method.
  }
  \resizebox{0.8\linewidth}{!}{%
    \begin{tabular}{l|ccccccc|ccccccc|c}
  \toprule[2pt]
  \rowcolor{tabtitle}
                                 &
  \multicolumn{7}{c|}{\textbf{CAD}}
                                 &
  \multicolumn{7}{c|}{\textbf{MoCA-Mask-TE}}
                                 &                                                                                                                                                                                                                                                                                                                                      \\
  \rowcolor{tabtitle}
  \multirow{-2}{*}{Model}        & S$_{m}~\uparrow$ & F$^{\omega}_{\beta}~\uparrow$ & MAE$~\downarrow$ & F$_{\beta}~\uparrow$ & E$_{m}~\uparrow$ & mDice$~\uparrow$ & mIoU$~\uparrow$ & S$_{m}~\uparrow$ & F$^{\omega}_{\beta}~\uparrow$ & MAE$~\downarrow$ & F$_{\beta}~\uparrow$ & E$_{m}~\uparrow$ & mDice$~\uparrow$ & mIoU$~\uparrow$ & \multirow{-2}{*}{ARG} \\
  \midrule[1pt]
  Ours$^{23}$                    & \best{0.757}     & \best{0.593}                  & \best{0.020}     & \best{0.631}         & \best{0.865}     & \best{0.599}     & \best{0.510}    & \best{0.734}     & \best{0.476}                  & \best{0.010}     & \best{0.497}         & \best{0.736}     & \best{0.497}     & \best{0.422}    & 0.00\%                     \\
  \midrule[0.5pt]
  w/ difference image            & 0.724            & 0.528                         & 0.023            & 0.569                & 0.761            & 0.526            & 0.441           & 0.693            & 0.401                         & 0.017            & 0.429                & 0.704            & 0.425            & 0.360           & $\downarrow$15.46\%        \\
  w/ optical flow                & 0.729            & 0.534                         & 0.022            & 0.573                & 0.765            & 0.532            & 0.450           & 0.697            & 0.409                         & 0.016            & 0.436                & 0.708            & 0.433            & 0.370           & $\downarrow$13.41\%        \\
  \midrule[0.5pt]
  w/o intra-frame self-attention & 0.741            & 0.552                         & 0.023            & 0.588                & 0.862            & 0.574            & 0.490           & 0.683            & 0.394                         & 0.013            & 0.412                & 0.684            & 0.413            & 0.352           & $\downarrow$10.79\%        \\
  w/o cross-frame cues diffusion & 0.736            & 0.546                         & 0.024            & 0.584                & 0.860            & 0.569            & 0.481           & 0.679            & 0.384                         & 0.013            & 0.404                & 0.681            & 0.405            & 0.340           & $\downarrow$12.17\%        \\
  w/o temporal shifting          & 0.742            & 0.571                         & 0.021            & 0.610                & 0.828            & 0.576            & 0.488           & 0.690            & 0.408                         & 0.013            & 0.426                & 0.692            & 0.427            & 0.366           & $\downarrow$8.88\%         \\
  \bottomrule[2pt]
\end{tabular}

  }
  \label{tab:videoablationstudy}
\end{table*}

\begin{table*}[t]
  \centering
  \caption{Comparisons of mixed and single scale input schemes. All models are based on the baseline model in Tab.~\ref{tab:ablationstudy}.}
  \resizebox{0.7\linewidth}{!}{%
    \begin{tabular}{cc|ll|ccccc|ccccc|c}
  \toprule[2pt]
  \rowcolor{tabtitle}
                                    &
                                    &
                                    &
                                    &
  \multicolumn{5}{c|}{\textbf{COD10K}}
                                    &
  \multicolumn{5}{c|}{\textbf{NC4K}}
                                    &                                                                                                                                                                                                                                                                                                                                                              \\
  \rowcolor{tabtitle}
  \multirow{-2}{*}{Input Scale}     & \multirow{-2}{*}{Strategy} & \multirow{-2}{*}{Param. (M)} & \multirow{-2}{*}{GFLOPs} & S$_{m}$ $\uparrow$ & F$^{\omega}_{\beta}$ $\uparrow$ & MAE $\downarrow$ & F$_{\beta}$ $\uparrow$ & E$_{m}$ $\uparrow$ & S$_{m}$ $\uparrow$ & F$^{\omega}_{\beta}$ $\uparrow$ & MAE $\downarrow$ & F$_{\beta}$ $\uparrow$ & E$_{m}$ $\uparrow$ & \multirow{-2}{*}{ARG} \\
  \midrule[1pt]
  $1.0\times$                       & \none                      & 25.216                       & 41.000                   & 0.826              & 0.677                           & 0.035            & 0.724                  & 0.903              & 0.850              & 0.756                           & 0.049            & 0.800                  & 0.912              & 0.00\%                \\
  $0.5\times$                       & \none                      & 25.216                       & 41.369                   & 0.761              & 0.552                           & 0.049            & 0.611                  & 0.860              & 0.810              & 0.686                           & 0.061            & 0.737                  & 0.882              & $\downarrow$13.63\%   \\
  $0.5\times, 1.0\times$            & Addition                   & 25.216                       & 48.987                   & 0.827              & 0.672                           & 0.034            & 0.722                  & 0.910              & 0.858              & 0.765                           & 0.046            & 0.805                  & 0.918              & $\uparrow$1.23\%      \\
  $0.5\times, 1.0\times$            & MHSIU                      & 25.952                       & 61.581                   & 0.833              & 0.689                           & 0.033            & 0.735                  & 0.911              & 0.857              & 0.768                           & 0.046            & 0.808                  & 0.916              & $\uparrow$2.07\%      \\
  $1.5\times$                       & \none                      & 25.216                       & 104.972                  & 0.850              & 0.717                           & 0.031            & 0.764                  & 0.921              & 0.856              & 0.769                           & 0.048            & 0.810                  & 0.913              & $\uparrow$3.36\%      \\
  $0.5\times, 1.5\times$            & Addition                   & 25.216                       & 120.565                  & 0.843              & 0.715                           & 0.032            & 0.760                  & 0.920              & 0.858              & 0.770                           & 0.047            & 0.812                  & 0.918              & $\uparrow$3.21\%      \\
  $0.5\times, 1.5\times$            & MHSIU                      & 26.137                       & 141.342                  & 0.853              & 0.719                           & 0.031            & 0.763                  & 0.922              & 0.862              & 0.774                           & 0.046            & 0.814                  & 0.918              & $\uparrow$4.07\%      \\
  $1.0\times, 1.5\times$            & Addition                   & 25.216                       & 112.589                  & 0.852              & 0.717                           & 0.031            & 0.761                  & 0.924              & 0.864              & 0.776                           & 0.046            & 0.814                  & 0.919              & $\uparrow$4.09\%      \\
  $1.0\times, 1.5\times$            & MHSIU                      & 25.952                       & 132.459                  & 0.856              & 0.722                           & 0.030            & 0.767                  & 0.927              & 0.866              & 0.780                           & 0.045            & 0.819                  & 0.920              & $\uparrow$4.96\%      \\
  $0.5\times, 1.0\times, 1.5\times$ & Addition                   & 25.216                       & 120.571                  & 0.849              & 0.719                           & 0.030            & 0.763                  & 0.923              & 0.863              & 0.782                           & 0.043            & 0.819                  & 0.921              & $\uparrow$5.15\%      \\
  \rowcolor{ours}
  $0.5\times, 1.0\times, 1.5\times$ & MHSIU                      & 26.530                       & 149.100                  & 0.856              & 0.728                           & 0.029            & 0.770                  & 0.923              & 0.870              & 0.789                           & 0.043            & 0.825                  & 0.924              & $\uparrow$6.03\%      \\
  $1.5\times, 1.5\times, 1.5\times$ & MHSIU                      & 26.530                       & 295.659                  & 0.853              & 0.719                           & 0.031            & 0.765                  & 0.924              & 0.864              & 0.779                           & 0.045            & 0.818                  & 0.918              & $\uparrow$4.47\%      \\
  \bottomrule[2pt]
\end{tabular}%

  }
  \label{tab:msi_scheme}
\end{table*}

\begin{table*}[t]
  \centering
  \caption{Comparisons of different down-sampling methods for the proposed \myMHSIU.}
  \resizebox{0.6\linewidth}{!}{%
    \begin{tabular}{l|*5{c}|*5{c}|c}
  \toprule[2pt]
  \rowcolor{tabtitle}
                     &
  \multicolumn{5}{c|}{\textbf{COD10K-TE}}
                     &
  \multicolumn{5}{c}{\textbf{NC4K}}
                     &                                                                                                                                                                                                                                                                \\
  \rowcolor{tabtitle}
  \multirow{-2}{*}{\textbf{Method}}
                     & $S_{m}~\uparrow$ & $F^{\omega}_{\beta}~\uparrow$ & MAE~$\downarrow$ & $F_{\beta}~\uparrow$ & $E_{m}~\uparrow$ & $S_{m}~\uparrow$ & $F^{\omega}_{\beta}~\uparrow$ & MAE~$\downarrow$ & $F_{\beta}~\uparrow$ & $E_{m}~\uparrow$ & \multirow{-2}{*}{\textbf{ARG}} \\
  \midrule[1pt]
  \rowcolor{ours}
  Ours               & 0.856            & 0.728                         & 0.029            & 0.770                & 0.923            & 0.870            & 0.789                         & 0.043            & 0.825                & 0.924            & 0.00\%                         \\
  w/ average-pooling & 0.853            & 0.724                         & 0.031            & 0.765                & 0.919            & 0.865            & 0.779                         & 0.046            & 0.821                & 0.920            & $\downarrow$1.86\%             \\
  w/ max-pooling     & 0.855            & 0.726                         & 0.030            & 0.767                & 0.920            & 0.867            & 0.783                         & 0.044            & 0.823                & 0.923            & $\downarrow$0.83\%             \\
  w/ bi-linear       & 0.852            & 0.722                         & 0.031            & 0.764                & 0.914            & 0.862            & 0.775                         & 0.047            & 0.818                & 0.920            & $\downarrow$2.32\%             \\
  w/ bi-cubic        & 0.853            & 0.723                         & 0.031            & 0.766                & 0.916            & 0.863            & 0.777                         & 0.046            & 0.819                & 0.921            & $\downarrow$1.96\%             \\
  \bottomrule[2pt]
\end{tabular}

  }
  \label{tab:downsample}
\end{table*}

\begin{table*}[t]
  \centering
  \caption{Comparisons of different increasing strategies of $\lambda$.
    $\lambda_{const}$: A constant value and it is set to $1$.
    $t$ and $T$: The current and total number of iterations, respectively.
    $\lambda_{min}$ and $\lambda_{max}$: The minimum and maximum values of $\lambda$, and they are set to $0$ and $1$ in our experiments.
    ``$\star_{t_{min} \rightarrow t_{max}}$'': The increasing interval in the iterations is $[t_{min}, t_{max}]$, where $\lambda = \lambda_{min}$ when $t \le t_{min}$, and $\lambda = \lambda_{max}$ when $t \ge t_{max}$.}
  \resizebox{0.8\linewidth}{!}{%
    \begin{tabular}{c|c|ccccc|ccccc}
  \toprule[2pt]
  \rowcolor{tabtitle}
                                   &                                                                                                                                    & \multicolumn{5}{c|}{\textbf{COD10K}} & \multicolumn{5}{c}{\textbf{NC4K}}                                                                                                                                                                                          \\
  \rowcolor{tabtitle}
  \multirow{-2}{*}{Strategy}       & \multirow{-2}{*}{$\lambda$}                                                                                                        & S$_{m}$ $\uparrow$                   & F$^{\omega}_{\beta}$ $\uparrow$    & MAE $\downarrow$ & F$_{\beta}$ $\uparrow$ & E$_{m}$ $\uparrow$ & S$_{m}$ $\uparrow$ & F$^{\omega}_{\beta}$ $\uparrow$ & MAE $\downarrow$ & F$_{\beta}$ $\uparrow$ & E$_{m}$ $\uparrow$ \\
  \midrule[1pt]
  Cosine$_{0 \rightarrow T}$       &                                                                                                                                    & \best{0.861}                         & \best{0.768}                       & 0.026            & \best{0.801}           & 0.925              & \best{0.874}       & \best{0.816}                    & \best{0.037}     & \best{0.846}           & \best{0.928}       \\
  Cosine$_{0.3T \rightarrow 0.7T}$ & \multirow{-2}{*}{$\lambda_{min} + \frac{1}{2}(1 - \cos(\frac{t - t_{min}}{t_{max} - t_{min}}\pi))(\lambda_{max} - \lambda_{min})$} & 0.858                                & 0.767                              & \best{0.025}     & 0.798                  & 0.925              & 0.870              & 0.813                           & \best{0.037}     & 0.841                  & 0.924              \\
  \midrule[1pt]
  Linear$_{0 \rightarrow T}$       &                                                                                                                                    & 0.859                                & 0.767                              & 0.026            & \best{0.801}           & 0.926              & 0.870              & 0.814                           & \best{0.037}     & 0.843                  & 0.927              \\
  Linear$_{0.3T \rightarrow 0.7T}$ & \multirow{-2}{*}{$\lambda_{min} + \frac{t - t_{min}}{t_{max} - t_{min}}(\lambda_{max} - \lambda_{min})$}                           & 0.856                                & 0.766                              & 0.026            & 0.799                  & 0.924              & 0.869              & 0.813                           & 0.038            & 0.843                  & 0.924              \\
  \midrule[1pt]
  Constant                         & $\lambda_{const}$                                                                                                                  & 0.859                                & 0.767                              & \best{0.025}     & 0.800                  & \best{0.928}       & 0.868              & 0.811                           & 0.039            & 0.839                  & 0.924              \\
  \bottomrule[2pt]
\end{tabular}%

  }
  \label{tab:exp_losscoef}
\end{table*}

\begin{table*}[t]
  \centering
  \caption{Different forms of the proposed \myUAL.
    Form 0 is our model without \myUAL.
    ``\none'': Unable to converge.
    Form 1.5 is used by default because of its balanced performance.
    Their curves are shown in Fig.~\ref{fig:loss_form}.
  }
  \resizebox{0.7\linewidth}{!}{%
    \begin{tabular}{c|c|c|*5{c}|*5{c}}
  \toprule[2pt]
  \rowcolor{tabtitle}
                                 &                                                                      &                                     & \multicolumn{5}{c|}{\textbf{COD10K-TE}} & \multicolumn{5}{c}{\textbf{NC4K}}                                                                                                                                                                              \\
  \rowcolor{tabtitle}
  \multirow{-2}{*}{\textbf{No.}} & \multirow{-2}{*}{\textbf{Form}}                                      & \multirow{-2}{*}{$\mathbf{\alpha}$} & $S_{m}~\uparrow$                        & $F^{\omega}_{\beta}~\uparrow$     & MAE~$\downarrow$ & $F_{\beta}~\uparrow$ & $E_{m}~\uparrow$ & $S_{m}~\uparrow$ & $F^{\omega}_{\beta}~\uparrow$ & MAE~$\downarrow$ & $F_{\beta}~\uparrow$ & $E_{m}~\uparrow$ \\
  \midrule[1pt]
  0                              & \no                                                                  & \no                                 & 0.861                                   & 0.734                             & 0.029            & 0.773              & 0.924              & 0.875            & 0.792                         & 0.042            & 0.826              & 0.925              \\
  \midrule[1pt]
  1.1, 1.2, 1.3                  &                                                                      & 1/8, 1/4, 1/2                       & \none                                   & \none                             & \none            & \none              & \none              & \none            & \none                         & \none            & \none              & \none              \\
  1.4                            &                                                                      & 1                                   & 0.859                                   & 0.762                             & \best{0.026}     & 0.795              & 0.923              & 0.870            & 0.810                         & 0.039            & 0.841              & 0.926              \\
  1.5                            &                                                                      & 2                                   & 0.861                                   & \best{0.768}                      & \best{0.026}     & 0.801              & 0.925              & 0.874            & \best{0.816}                  & \best{0.037}     & \best{0.846}       & \best{0.928}       \\
  1.6                            &                                                                      & 4                                   & 0.857                                   & \best{0.768}                      & 0.027            & \best{0.802}       & 0.923              & 0.868            & 0.815                         & 0.038            & 0.844              & 0.925              \\
  1.7                            & \multirow{-5}{*}{$\Phi_{pow}^{\alpha}(x)=1-|2x-1|^{\alpha}$}         & 8                                   & 0.859                                   & 0.767                             & \best{0.026}     & \best{0.802}       & 0.924              & 0.867            & 0.814                         & 0.038            & 0.843              & 0.923              \\
  \midrule[1pt]
  2.1                            &                                                                      & 1/8                                 & \best{0.866}                            & 0.741                             & 0.027            & 0.782              & \best{0.933}       & 0.875            & 0.794                         & 0.041            & 0.828              & 0.926              \\
  2.2                            &                                                                      & 1/4                                 & 0.862                                   & 0.736                             & 0.028            & 0.777              & 0.928              & \best{0.876}     & 0.794                         & 0.041            & 0.828              & 0.926              \\
  2.3                            &                                                                      & 1/2                                 & 0.863                                   & 0.742                             & 0.028            & 0.781              & 0.930              & \best{0.876}     & 0.798                         & 0.040            & 0.831              & 0.927              \\
  2.4                            &                                                                      & 1                                   & 0.863                                   & 0.753                             & \best{0.026}     & 0.790              & 0.927              & 0.873            & 0.802                         & 0.039            & 0.834              & 0.925              \\
  2.5                            &                                                                      & 2                                   & 0.862                                   & 0.762                             & \best{0.026}     & 0.797              & 0.927              & 0.873            & 0.810                         & 0.038            & 0.841              & \best{0.928}       \\
  2.6                            &                                                                      & 4                                   & 0.861                                   & 0.752                             & 0.027            & 0.790              & 0.925              & 0.872            & 0.803                         & 0.040            & 0.835              & 0.925              \\
  2.7                            & \multirow{-7}{*}{$\Phi_{exp}^{\alpha}(x)=e^{-(\alpha (x-0.5))^{2}}$} & 8                                   & 0.861                                   & 0.737                             & 0.028            & 0.778              & 0.929              & 0.874            & 0.793                         & 0.041            & 0.827              & 0.926              \\
  \midrule[1pt]
  3                              & BCE w/ $\omega = 1+\Phi_{pow}^{2}(x)$                                & 2                                   & 0.861                                   & 0.737                             & 0.028            & 0.777              & 0.928              & 0.873            & 0.791                         & 0.041            & 0.825              & 0.926              \\
  \bottomrule[2pt]
\end{tabular}

  }
  \label{tab:loss_form}
\end{table*}

\parhead{\myMHSIU}
As shown in Fig.~\ref{fig:siuattn}, the multi-head paradigm of the \myMHSIU further enriches the spatial pattern of attention to fine-grained information in different scale spaces.
We list the efficiency and performance of different settings in Tab.~\ref{tab:ablationstudy}.
The comparison shows that an increase in the number of heads improves computing efficiency and even performance.
The best performance appears when the number of heads is 4, which is also the choice in other experiments.

The above experiments reflect from the side that the increase of model complexity does not necessarily lead to the improvement of performance, which also depends on more reasonable structural designs.

\subsubsection{Temporal Conditional Computation}

\parhead{Number of Frames}
Our model introduces the temporal conditional computation, \ie~\myrouting, which unifies the feature processing pipelines of image and video tasks into the same framework.
Tab.~\ref{tab:vcod-sota} shows the performance corresponding to different versions of the model on the video COD task.
Since there is no difference in the time dimension when only a single frame is input, the corresponding model only uses static processing nodes.
As more frames are fed, adaptively activated network nodes by differential information for the inter-frame interaction further improve the performance on video COD.

\parhead{Motion Modeling}
In Tab.~\ref{tab:videoablationstudy}, we also introduce two common variants concatenated with motion features from the difference map and optical flow.
Both result in an increased computational burden beyond the visual model due to obtaining motion information, and the performance drops are 15.46\% and 13.41\% respectively when facing camouflaged objects.

\parhead{Finer-grained Operations}
Moreover, in Tab.~\ref{tab:videoablationstudy}, the performance degradation caused by removing the finer-grained operations in the temporal computation of the \myRGPU, further validates the important role they each play.
Note that if removing the key \texttt{shift} operation in the temporal conditional computation, not only the overall performance will degrade, but also the temporal bypass will lose its dynamics.

\subsubsection{Mixed-scale Input Scheme}
\label{sec:msi}

\parhead{Mixed Scales}
Our model is designed to mimic the behavior of ``zooming in and out''.
The feature expression is enriched by combining the scale-specific information from different scales.
In Fig.~\ref{fig:featvisualablation}, the intermediate features from the deep \myMHSIU modules show that our mixed-scale scheme plays a positive and important role in locating the camouflaged object.
To analyze the role of different scales, we summarize the performance of different combination forms in Tab.~\ref{tab:msi_scheme}.
Our scheme performs better than the single-scale one and simply mixed one.
And it also surpasses the single-scale form integrating three $1.5 \times$ scales in the accuracy and efficiency, which reflects that multi-scale cues are key to the value of \myMHSIU.
This verifies the rationality of such a design for the COD task.

\parhead{Down-sampling}
In the \myMHSIU, we down-sample the high-resolution feature $f^{1.5}_{i}$ to supplement effective and diverse responses for camouflaged objects to $f^{1.0}_{i}$.
To select a suitable down-sampling method, we analyze several common techniques in Tab.~\ref{tab:downsample}, including average-pooling, max-pooling, bi-linear interpolation, and bi-cubic interpolation.
As can be seen from Tab.~\ref{tab:downsample}, ours based on max-pooling and average-pooling achieves better performance.
So this setting is used by default in our framework.

\subsubsection{Forms of Loss Function}
\label{sec:loss_form}

\parhead{Options of Setting $\lambda$}
We compare three strategies and the results are listed in Tab.~\ref{tab:exp_losscoef}, in which the increasing cosine strategy achieves the best performance.
This may be due to the advantage of its smooth change process.
This smooth intensity warm-up strategy of \myUAL motivates the model to take advantage of \myUAL in improving the learning process and to mitigate the possible negative interference of \myUAL on BCE due to the lower accuracy of the model during the early stage of training.

\parhead{Forms of \myUAL}
Different forms of \myUAL are listed in Tab.~\ref{tab:loss_form} and the corresponding curves are illustrated in Fig.~\ref{fig:loss_form}.
As can be seen, Form 1.5 has a more balanced performance.
Also, it is worth noting that, when $\Delta$ approaches 1, the form, which can maintain a larger gradient, will obtain better performance in terms of F$^{\omega}_{\beta}$, MAE, and F$_{\beta}$.

\begin{figure}[t]
  \centering
  \subfloat%
  {%
    \centering
    \includegraphics[width=\linewidth]{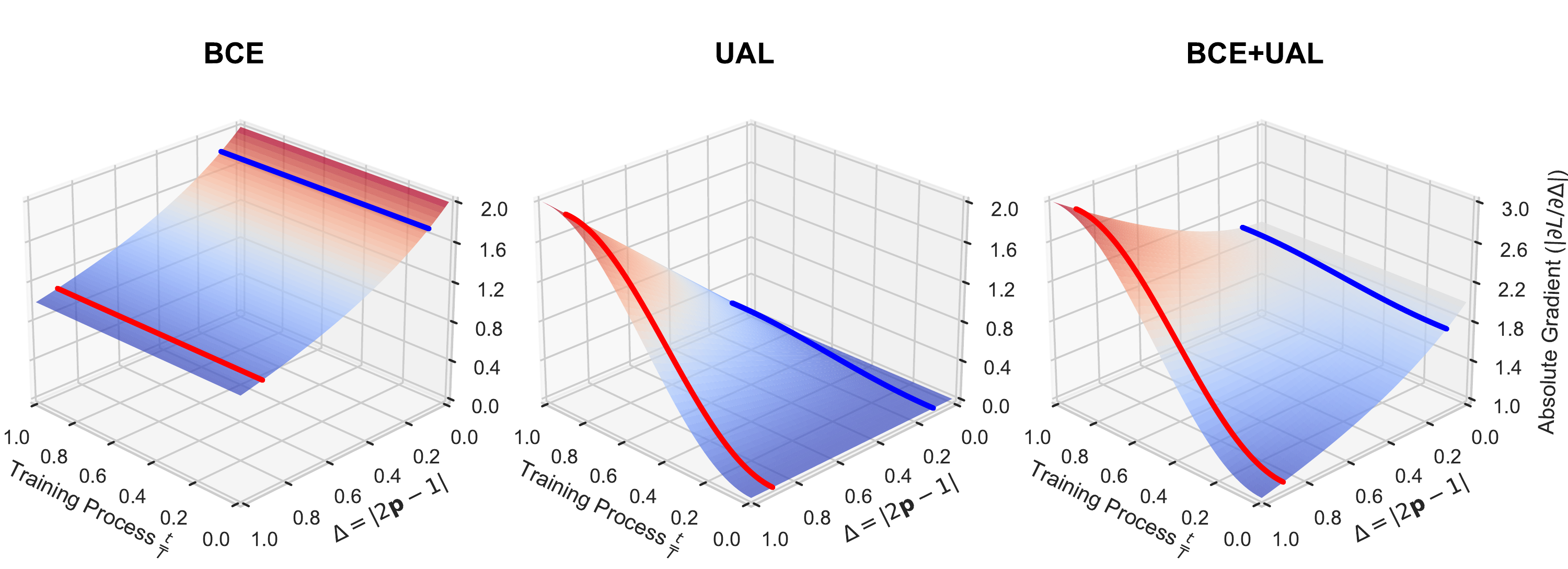}
    \label{fig:gradsurf}
  }
  \\
  \subfloat%
  {%
    \centering
    \includegraphics[width=\linewidth]{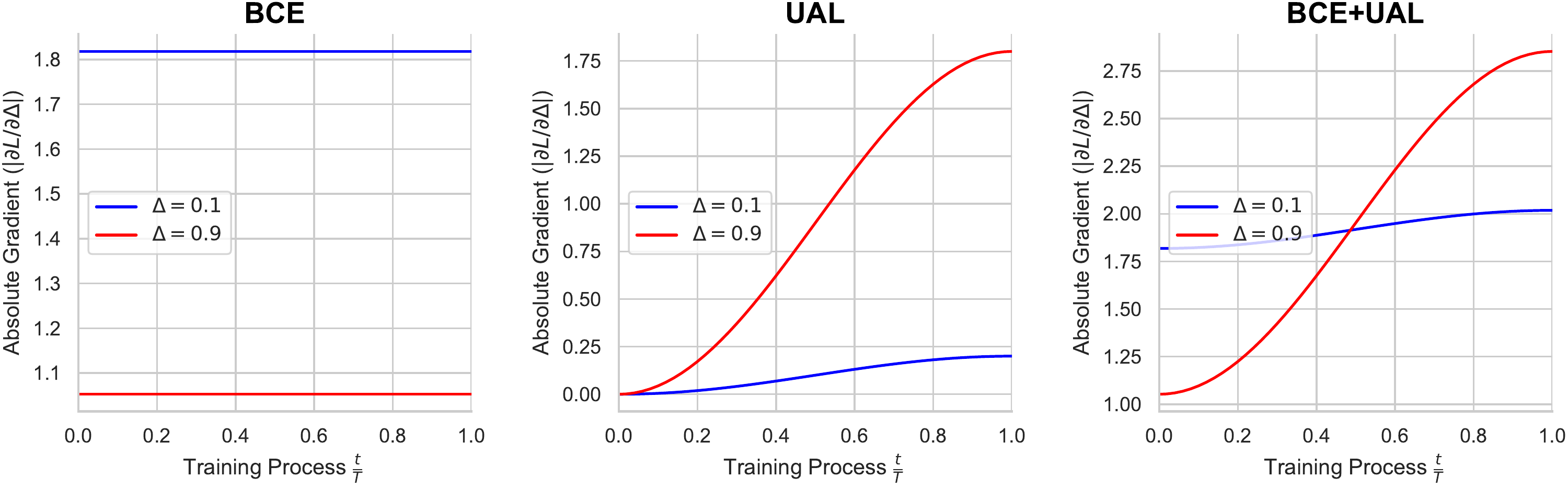}
    \label{fig:gradcurve}
  }
  \caption{Gradient landscapes (\textit{Top}) and curves (\textit{Bottom}) of the binary cross entropy (BCE) loss and our \myual (\myUAL) scheduled by cosine with respect to the measure $\Delta = |2\mathbf{p} - 1|$ as in Equ.~\ref{equ:ual} and the training progress $\frac{t}{T}$, where $t$ and $T$ denote the current and overall iteration steps.}
  \label{fig:lossgrad}
  \vspace{-1em}
\end{figure}

\parhead{Gradient of Loss}
Further, from the gradient of the loss function illustrated in Fig.~\ref{fig:lossgrad}, there is a good collaboration between the proposed \myUAL and BCE.
In the early training phase, the \myUAL is suppressed and the gradient is dominated by the BCE, thus facilitating a more stable optimization.
However, in the later stage, the gradient of \myUAL is inversely proportional to $\Delta$ representing the certainty of the prediction.
Such a design enables \myUAL to speed up updating simple samples.
At the same time, it does not rashly adjust difficult samples, because they need more guidance from the ground truth.

\begin{figure}[t]
  \centering
  \includegraphics[width=\linewidth]{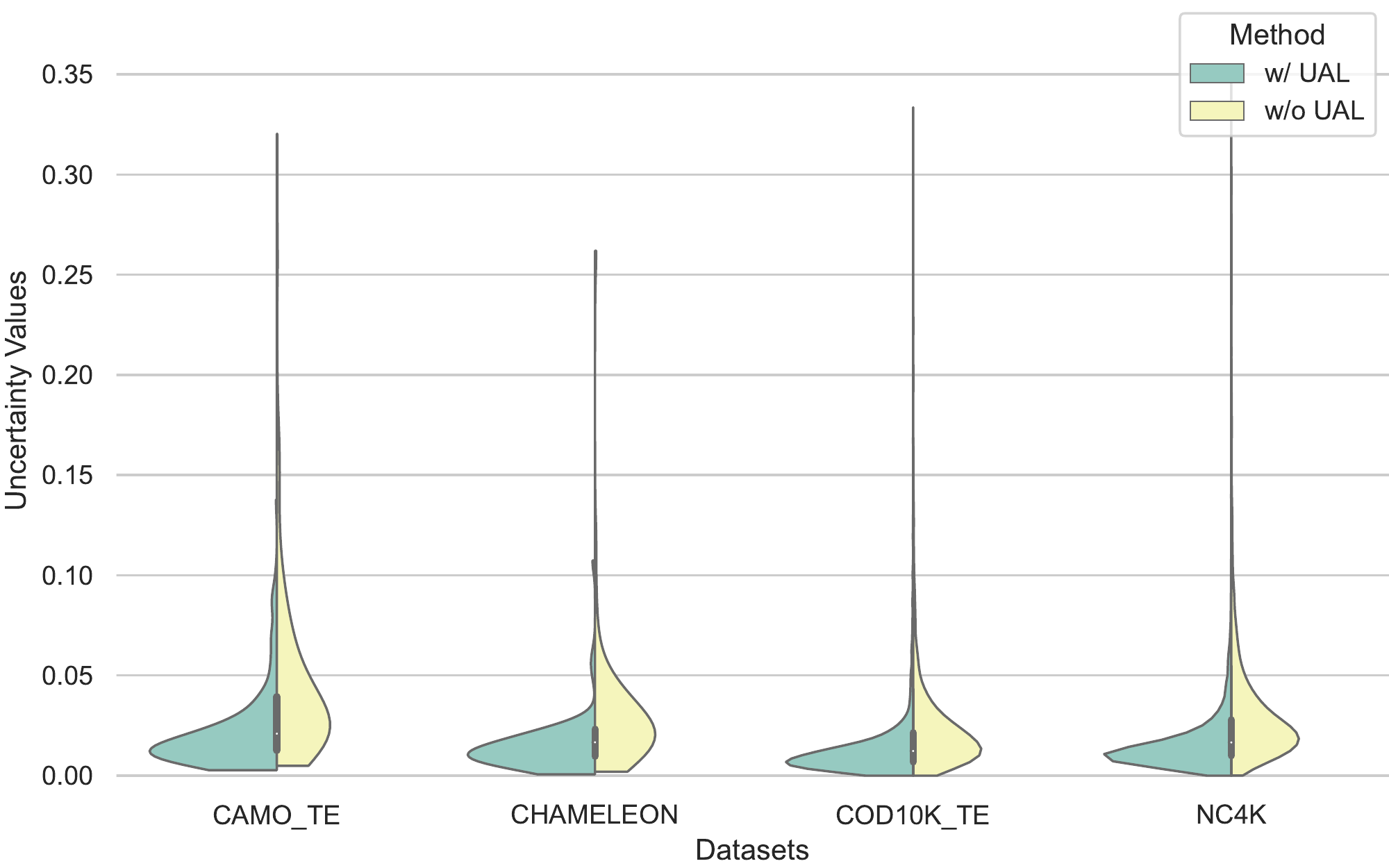}
  \caption{
    Visual comparison based on the violin plot for the proposed model with/without \myUAL.
    It counts the uncertainty measure of the prediction results for four image COD datasets, CAMO-TE~\cite{CAMO}, CHAMELEON~\cite{CHAMELEON}, COD10K~\cite{COD10K}, and NC4K~\cite{SLSR}.
    Better performance requires that the overall uncertainty distribution tends more toward zero.
    It is best to zoom in for more details.
  }
  \label{fig:hist}
  \vspace{-1em}
\end{figure}

\subsubsection{Effectiveness of \myUAL}
\label{sec:ual}

We also introduce some recent GT-free losses including $L_{ADB}$~\cite{A2Sv1}, $L_{CSD}$~\cite{A2Sv2}, $L_{BTM}$~\cite{A2Sv2}, and $L_{MS}$~\cite{A2Sv2} in Tab.~\ref{tab:ablationstudy}.
Unfortunately, some of them (\textit{i.e.}, $L_{BTM}$ and the combined version $L_{CSD}+L_{BTM}+L_{MS}$) fail to converge due to differences in applicable scenarios and training settings.
But overall, our $L_{UAL}$ shows a better performance.
Besides, Fig.~\ref{fig:visualation} and Fig.~\ref{fig:featvisualablation} intuitively show that the \myUAL greatly reduces the ambiguity caused by the interference from the background.
We also visualize the uncertainty measure, \ie~\myUAL, of all results on four image COD datasets in Fig.~\ref{fig:hist}.
The peaks of the model ``w/o \myUAL'' are flatter and more upward, which means that there are more visually blurred/uncertain predictions.
Besides, in the corresponding feature visualization in Fig.~\ref{fig:featvisualablation}, there is clear background interference in ``w/o \myUAL'' due to the complex scenarios and blurred edges, which are extremely prone to yield false positive predictions.
However, when \myUAL is introduced, it can be seen in Fig.~\ref{fig:hist} that the peaks of ``w/ \myUAL'' are sharper than the one of ``w/o \myUAL'', that is, most pixel values approach two extremes and they have higher confidence.
And the feature maps in Fig.~\ref{fig:featvisualablation} become more discriminative and present a more compact and complete response in the regions of camouflaged objects.

\section{Conclusion}

In this paper, we propose \myModel by imitating the behavior of human beings to zoom in and out on images and videos.
This process actually considers the differentiated expressions about the scene from different scales, which helps to improve the understanding and judgment of camouflaged objects.
We first filter and aggregate scale-specific features through the scale merging subnetwork to enhance feature representation.
Next, in the hierarchical difference propagation decoder, the strategies of grouping, mixing, and fusion further mine the mixed-scale semantics.
Lastly, we introduce the \myual to penalize the ambiguity of the prediction.
And, to the best of our knowledge, we are the first to unify the pipelines of image and video COD into a single framework.
Extensive experiments verify the effectiveness of the proposed method in both the image and video COD tasks with superior performance to existing state-of-the-art methods.

\parhead{Limitations}
While our \myModel offers a potent and efficacious approach to the COD challenge, opportunities for further investigation and enhancement remain.
In our current design, a shared feature extraction architecture actively gathers complementary information across a range of scales from the image pyramid, emulating the dynamic process of zooming in and out to capture detail.
However, unlike our explicit approach, humans typically assimilate and internalize information effortlessly during the learning process.
Despite being highly accurate for both image and video COD tasks, our scale-independent model inherently demands extra computational resources during inference.

\parhead{Future Work}
How to further optimize the computational burden of the feature extraction component remains an important and open question.
Techniques such as more compact architectural designs and scale knowledge distillation~\cite{ScaleKD,MSAlignedKD} are directions worth exploring.

  {\small
    \bibliographystyle{plainnat}
    \bibliography{IEEEabrv,Mybibabrv}}

\end{document}